\documentclass[lettersize,journal]{IEEEtran}
\usepackage{amsmath,amsfonts}
\usepackage{algorithmic}
\usepackage{algorithm}
\usepackage{array}
\usepackage[caption=false,font=normalsize,labelfont=sf,textfont=sf]{subfig}
\usepackage{textcomp}
\usepackage{stfloats}
\usepackage{url}
\usepackage{verbatim}
\usepackage{graphicx}
\usepackage{cite}
\usepackage{booktabs}
\usepackage{multirow}
\usepackage[table,dvipsnames]{xcolor}
\usepackage{bm}
\usepackage{makecell}
\usepackage{pifont}       
\usepackage{bbding}       %
\usepackage{xcolor}
\usepackage[colorlinks=true,
            linkcolor=blue,
            citecolor=blue,
            urlcolor=blue]{hyperref}
\usepackage[switch]{lineno}
\usepackage{orcidlink}
\newcommand{\eg}{\emph{e.g.}}

\newcommand{\etal}{\emph{et al.}}

\definecolor{table_color}{RGB}{240, 240, 240}
\definecolor{table_color2}{RGB}{240, 240, 240}
\definecolor{attr_color}{RGB}{0, 0, 0}
\definecolor{attr_color2}{RGB}{0, 0, 0}

\definecolor{improve_color}{RGB}{0, 222, 0}

\begin{document}

\title{NEXT: Multi-Grained Mixture of Experts via Text-Modulation \\ for Multi-Modal Object Re-Identification}

\author{Shihao Li$^{\orcidlink{0009-0001-0923-3965}}$, Huaibo Huang$^{\orcidlink{0000-0001-5866-2283}}$, Junxian Duan$^{\orcidlink{0000-0002-0218-6924}}$, Aihua Zheng\(^{*\orcidlink{0000-0002-9820-4743}}\), Jin Tang$^{\orcidlink{0000-0001-8375-3590}}$, and Jixin Ma$^{\orcidlink{0000-0001-7458-7412}}$
\thanks{
This research is supported in part by the National Natural Science Foundation of China under Grants 62372003, and the Natural Science Foundation of Anhui Province under Grants 2308085Y40. (\(^*\)The corresponding author is Aihua Zheng.)
}
\thanks{
A. Zheng and S. Li are with the State Key Laboratory of Opto-Electronic Information Acquisition and Protection Technology, Anhui Provincial Key Laboratory of Multimodal Cognitive Computation, School of Artificial Intelligence, Anhui University, Hefei, 230601, China
(e-mail: ahzheng214@foxmail.com; shli0603@foxmail.com).
}
\thanks{
H. Huang and J. Duan are with the State Key Laboratory of Multimodal Artificial Intelligence Systems and the New Laboratory of Pattern Recognition, CASIA, Beijing, 100190, China, and also with the School of Artificial Intelligence, University of Chinese Academy of Sciences, Beijing, 100190, China
(e-mail: huaibo.huang@cripac.ia.ac.cn; junxian.duan@ia.ac.cn).
}
\thanks{
J. Tang is with Anhui Provincial Key Laboratory of Multimodal Cognitive Computation, School of Computer Science and Technology, Anhui University, Hefei, 230601, China 
(e-mail: tangjin@ahu.edu.cn).
}
\thanks{
J. Ma is with School of Computing and Mathematical Sciences, University of Greenwich, London SE10 9LS, UK
(e-mail: j.ma@greenwich.ac.uk).
}

}

\markboth{Journal of \LaTeX\ Class Files,~Vol.~14, No.~8, August~2021}%
{Shell \MakeLowercase{\textit{et al.}}: A Sample Article Using IEEEtran.cls for IEEE Journals}


\maketitle

\begin{abstract}
Multi-modal object Re-IDentification (ReID) aims to obtain complete identity features across heterogeneous modalities.
However, most existing methods rely on implicit feature fusion modules, making it difficult to model fine-grained recognition patterns under various challenges in real world.
Benefiting from the powerful Multi-modal Large Language Models (MLLMs), the object appearances are effectively translated into descriptive captions.
In this paper, we propose a reliable caption generation pipeline based on attribute confidence, which significantly reduces the unknown recognition rate of MLLMs and improves the quality of generated text.
Additionally, to model diverse identity patterns, we propose a novel ReID framework, named \textbf{NEXT}, the Multi-grained Mixture of Experts via Text-Modulation for Multi-modal Object Re-Identification. 
Specifically, we decouple the recognition problem into semantic and structural branches to separately capture fine-grained appearance features and coarse-grained structure features. 
For semantic recognition, we first propose a Text-Modulated Semantic Experts (TMSE), which randomly samples high-quality captions to modulate experts capturing semantic features and mining inter-modality complementary cues.
Second, to recognize structure features, we propose a Context-Shared Structure Experts (CSSE), which focuses on the holistic object structure and maintains identity structural consistency via a soft routing mechanism.
Finally, we propose a Multi-Grained Features Aggregation (MGFA), which adopts a unified fusion strategy to effectively integrate multi-grained expert features into the final identity representations.
Extensive experiments on two public person datasets and three vehicle datasets demonstrate the effectiveness of our method, showing that it significantly outperforms existing state-of-the-art methods.
\end{abstract}

\begin{figure}[t!]
\centering
\includegraphics[width=1.0\linewidth]{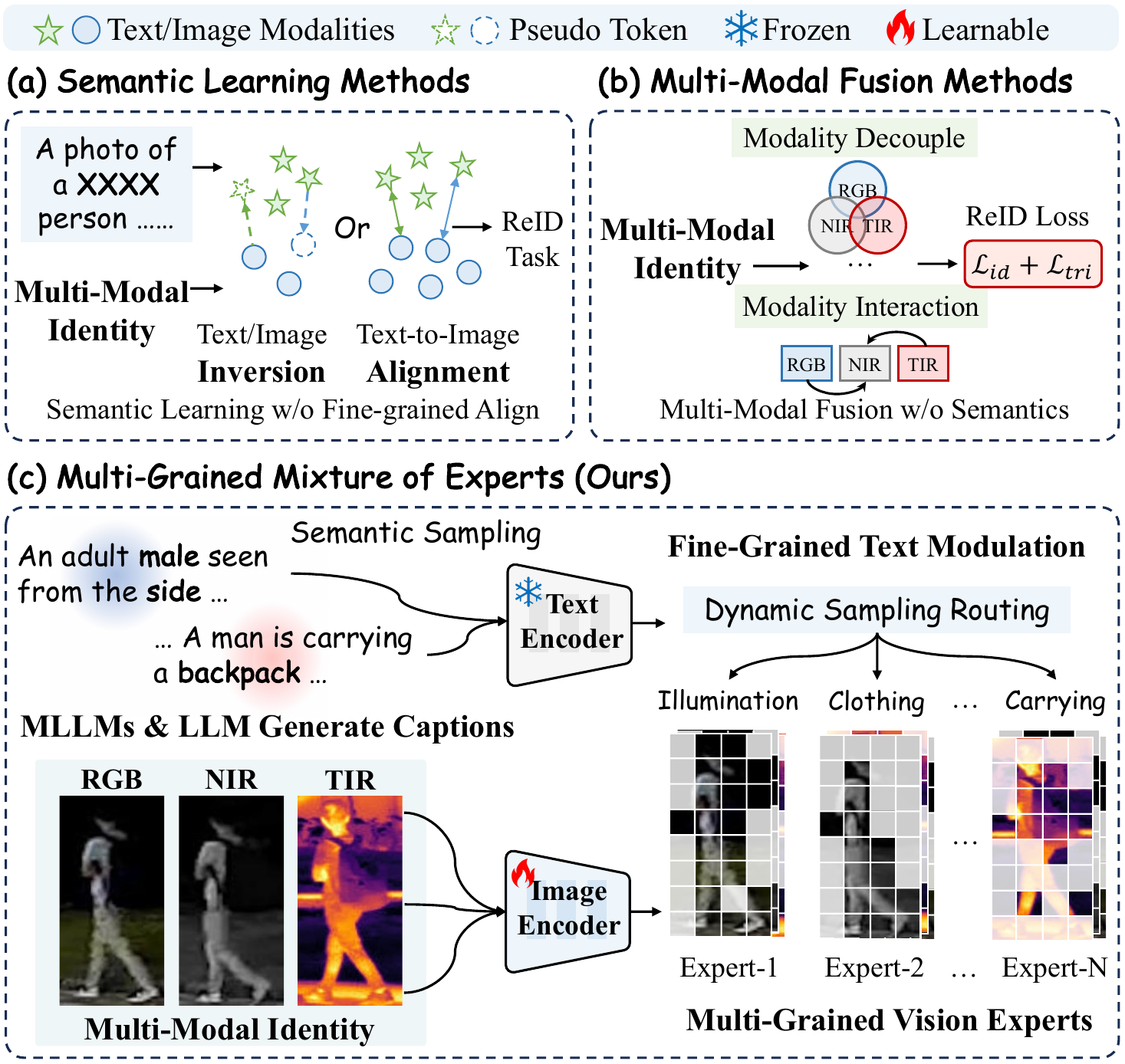}
\caption{Comparison with existing methods:
(a) Previous semantic learning methods capture identity information via text/image prompt inversion or alignment strategies, which fail to achieve fine-grained alignment.
(b) Existing multi-modal fusion methods fuse modality features through implicit fusion modules such as modal decoupling and interaction strategies, while the fusion process still lacks explicit semantic guidance.
(c) Our proposed method modulates visual modalities via diverse text captions and decouples appearance and structure into multi-grained experts.
}
\label{fig:motivation}
\end{figure}

\section{Introduction}
\IEEEPARstart{O}BJECT Re-IDentification (ReID) aims to construct discriminative identity representations by capturing object characteristics such as appearance, clothing, body shape, and posture~\cite{DBLP:journals/tcsv/LengYT20,ye2021deep,DBLP:conf/iccv/He0WW0021,DBLP:conf/cvpr/0004GLL019}. 
However, in real world scenarios, variations in illumination, weather, and occlusion~\cite{DBLP:conf/eccv/SunZYTW18,10098634,DBLP:conf/aaai/ChenCYYDK22,DSFAD,CSANet} hinder the performance of classical ReID methods. 
To address these challenges, multi-modal object ReID~\cite{DBLP:conf/aaai/Li0ZZ020,DBLP:journals/inffus/ZhengZMLTM23,DBLP:conf/aaai/ZhengWCLT21,DBLP:journals/inffus/ZhengMSWLT25} has attracted increasing research interest in recent years.
Benefiting from the complementary advantages of diverse spectral modalities, the identity features can be comprehensively captured.
Despite this, intra-modal noise and inter-modal heterogeneity remain the key limitations for effective identity representation fusion.
To overcome these challenges, existing methods focus on implicit fusion methods such as identity-level feature interaction~\cite{DBLP:conf/aaai/WangLZLTL24,10955143,Wang2024MambaPro}, fusion feature decoupling~\cite{wang2024demo,yu2024representation}, and frequency-domain interactions~\cite{yang2025tienet,zhang2024magic}. 
However, explicit semantic-aware fusion~\cite{wang2025idea,li2025icpl} of multi-modal identity features remains underexplored.

\begin{figure}[t]
\centering
\includegraphics[width=1.0\linewidth]{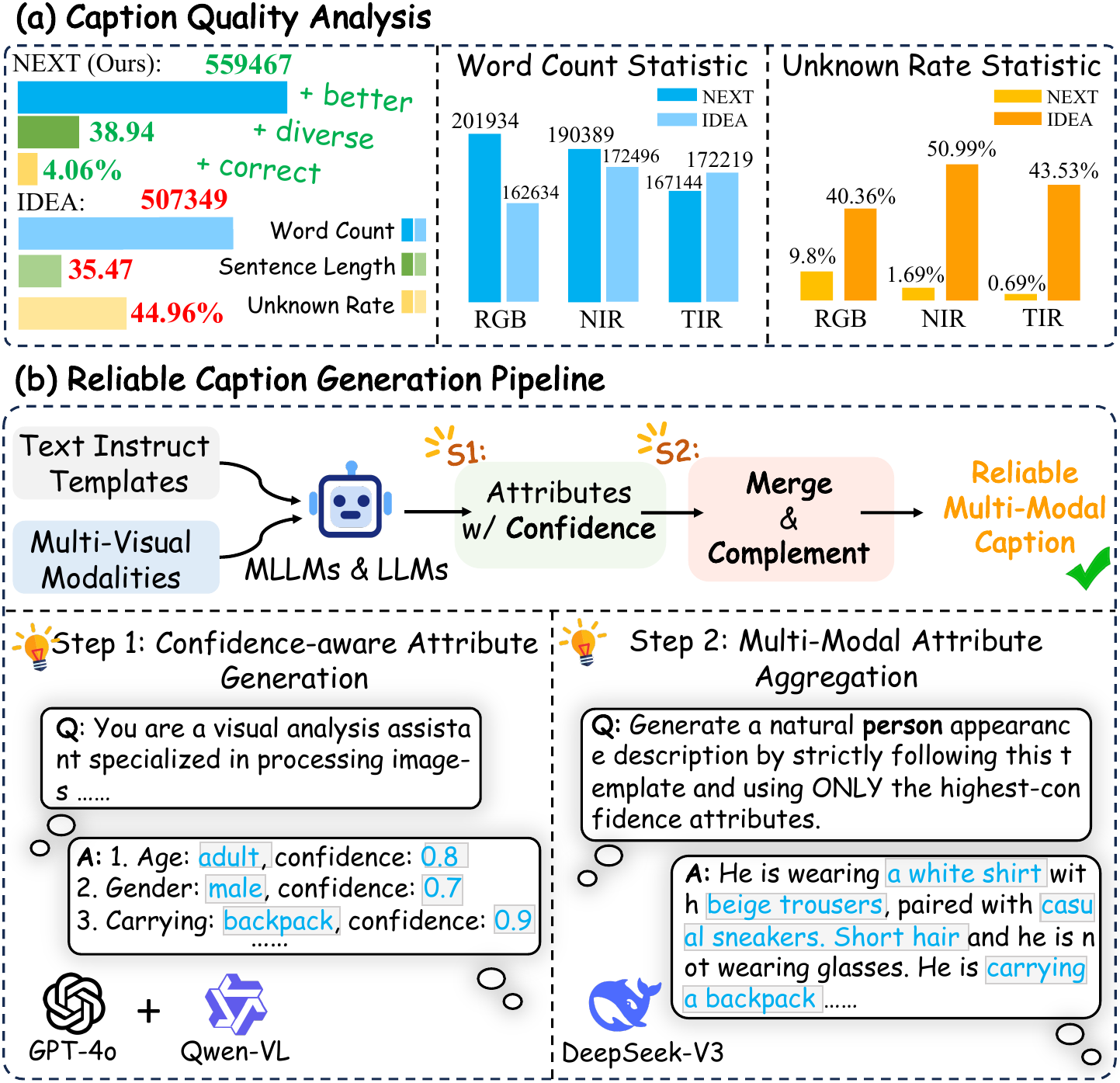}
\caption{
Caption analysis and generation pipeline. We first report the word count and the rate of captions containing `\textit{none},' `\textit{unknown},' or `\textit{unclear}' words to assess generation quality, and then show the reliable caption generation pipeline with MLLMs.
}
\label{fig:motivation_2}
\end{figure}

Recent studies in vision-language models~\cite{hurst2024gpt4o,bai2023qwenvl,DBLP:conf/nips/LiuLWL23a,DBLP:conf/icml/RadfordKHRGASAM21,DBLP:conf/icml/0001LXH22} have advanced visual understanding tasks~\cite{DBLP:conf/aaai/LiSL23,DBLP:conf/cvpr/YangWW0G024}.
Pioneering methods, such as IDEA~\cite{wang2025idea}, TVI-LFM~\cite{hu2024empowering}, and MP-ReID~\cite{DBLP:conf/aaai/ZhaiZH0JC24}, integrate MLLMs into ReID tasks by generating identity-relevant captions, thereby extending identity representations into the textual modality.
However, low-quality spectral noise and style discrepancies introduce semantic noise and omissions into the text generation process of MLLMs.
Information bias during long-context text generation serves as a major cause of hallucinations~\cite{fu2024mitigating} in MLLMs, with longer texts increasing the likelihood of such errors.
Additionally, guiding MLLMs to reason the fine-grained attributes across visible and infrared spectra is a highly labor-intensive process.
To tackle this, we propose a reliable caption generation framework, as shown in Fig.~\ref{fig:motivation_2}\textcolor{blue}{(b)}. 
Specifically, we decompose the caption generation process into two steps: confidence-aware attribute generation and multi-modal attribute aggregation.
By enforcing the model to output a confidence score for each attribute, we encourage the MLLMs to perform fine-grained evaluation for each attribute. 
Based on the confidence scores, we quantify the attribute importance across different modalities, enabling us to complement missing semantic information through multi-modal attribute sets. 

The high-quality captions effectively guide the model to focus on identity-relevant representations. 
Based on this, we propose a multi-grained mixture of experts framework named NEXT, as shown in Fig.~\ref{fig:motivation}\textcolor{blue}{(c)}, which models the identity recognition process through semantic-sampling and structure-aware experts.
The framework consists of three main components: Text-Modulated Semantic Experts (TMSE), Context-Shared Structure Experts (CSSE), and Multi-Grained Features Aggregation (MGFA).
Specifically, TMSE samples part-level fine-grained features within each modality. 
The text modulation mechanism encourages experts to focus on semantically relevant regions and capture intra-modality semantic identity features.
Meanwhile, CSSE perceives inter-modality coarse-grained features of the object identity through a soft routing mechanism shared across modalities, which maintains the integrity of structural features. 
Finally, MGFA unifies the features of multi-grained experts into a comprehensive identity representation.
Extensive experiments on five multi-modal object ReID benchmarks demonstrate the effectiveness of our proposed approach.

In summary, our contributions are as follows:
\begin{itemize}
\item To obtain accurate object appearance descriptions, we propose a reliable caption generation pipeline that leverages attribute confidence scores to effectively harness MLLMs in producing reliable high-quality captions.
\item To model diverse identity patterns, we propose the Text-Modulated Semantic Experts (TMSE) and the Context-Shared Structure Experts (CSSE) to decouple the recognition problem into fine-grained semantic and coarse-grained structural branches. The Multi-Grained Features Aggregation (MGFA) is proposed to integrate multi-grained experts into a unified identity representation.
\item Extensive experiments are conducted on five public benchmarks to validate the effectiveness of our method. The results demonstrate that the proposed method significantly outperforms the state-of-the-art methods.
\end{itemize}

\section{Related Work}

\subsection{Multi-Modal Object Re-Identification}
Multi-modal object ReID leverages the complementary imaging advantages of multi-spectra to enable robust identity recognition under adverse conditions. 
However, intra-modal noise and inter-modal spectral heterogeneity remain significant challenges. 
To fuse the spectral modalities, Wang~\etal~\cite{DBLP:conf/aaai/WangLZLTL24} propose the TOP-ReID which introduces a modality permutation and reconstruction mechanism to align multi-modal representations. 
Zhang~\etal~\cite{10955143} propose a prompt-based token selection and fusion strategy to effectively integrate multi-modal features and suppress background interference. 
Wang~\etal~\cite{Wang2024MambaPro} propose the MambaPro, which presents a selective state-space fusion approach based on Mamba~\cite{mamba} to aggregate intra- and inter-modal features. 
Zheng~\etal~\cite{DBLP:journals/inffus/ZhengMSWLT25} propose the FACENet which utilizes illumination priors to enhance degraded modalities. 
Wang~\etal~\cite{zhang2024magic} adopt frequency- and feature-based selection mechanisms to filter out background and low-quality noise. 
Yang~\etal~\cite{yang2025tienet} propose the TIENet which introduces an amplitude-guided phase learning strategy for frequency-domain enhancement. 
Li~\etal~\cite{li2025icpl} propose the ICPL-ReID which applies an identity prototype-based conditional prompt learning method to guide semantic learning across modalities. 
Wang~\etal~\cite{wang2025idea} inverts textual features into the visual space to guide model learning, and captures discriminative features across multi-modal via deformable aggregation. 
Despite these advancements, existing methods still lack a deep exploration of identity semantics and intrinsic structures. 
To bridge this gap, we propose a multi-grained mixture of experts via text-modulation that fuses features from different modalities through semantic and structural perspectives, aiming to enhance multi-modal identity representation based on explicit semantics.

\subsection{Semantic Learning in Re-Identification}
The large-scale Vision-Language foundation Models (VLM), such as CLIP~\cite{DBLP:conf/icml/RadfordKHRGASAM21}, exhibit powerful visual perception and semantic understanding capabilities. 
Researchers employ the learnable prompt~\cite{DBLP:journals/ijcv/ZhouYLL22,zhou2022conditional} to effectively adapt these models to classification and retrieval recognition tasks. 
In ReID, methods like CLIP-ReID~\cite{DBLP:conf/aaai/LiSL23} and PromptSG~\cite{DBLP:conf/cvpr/YangWW0G024} utilize learnable semantic prompts to transfer the CLIP~\cite{DBLP:conf/icml/RadfordKHRGASAM21} model to capture obejct identity semantic information~\cite{DBLP:journals/tifs/HeCWLWJD24}. 
With the rise of MLLMs~\cite{hurst2024gpt4o,bai2023qwenvl,DBLP:conf/nips/LiuLWL23a}, numerous studies demonstrate the effectiveness of MLLMs~\cite{DBLP:conf/nips/Dai0LTZW0FH23,DBLP:journals/corr/abs-2412-01720} in generating high-quality text and robust generalization in vision-language generation. 
Current approaches, such as MP-ReID~\cite{DBLP:conf/aaai/ZhaiZH0JC24}, TVI-LFM~\cite{hu2024empowering}, and IDEA~\cite{wang2025idea}, exploit MLLMs to generate identity-level textual descriptions from existing ReID datasets. 
However, these methods still lack reliable text generation under multi-modal conditions. 
Unlike existing methods~\cite{hu2024empowering,wang2025idea}, we further quantify each generated attribute and propose a confidence-aware attribute generation strategy. 
We incorporate a multi-modal merging and complementation mechanism to achieve high-quality and reliable caption generation. 
Additionally, we introduce the NEXT framework, which leverages semantic experts to extract fine-grained identity features, aiming to tackle diverse semantic challenges in multi-modal scenarios.

\subsection{Mixture of Experts for Multi-Modal Learning}
The Mixture-of-Experts (MoE)~\cite{moe1,moe2,DBLP:journals/neco/JacobsJNH91} paradigm decomposes a large and complex deep model into lightweight expert networks and a routing module. 
By intelligently routing inputs to different experts, the model enables flexible adaptation to diverse and complex tasks and environments~\cite{softmoe,sparsemoe}.
Recently, the MoE framework has been increasingly adopted in multi-modal learning~\cite{DBLP:conf/iclr/0008STWS25,flexemoe,DBLP:conf/nips/HanNHHS24}.
Yun~\etal~\cite{flexemoe} propose the Flex-MoE, which introduces a flexible framework that effectively incorporates arbitrary modality combinations and addresses the missing modality scenario across various domains.
Fang~\etal~\cite{emoe} recognize the modality importance varying across samples in fusion, and propose the EMoE which uses the predictive information of each modality to guide the fused features, preserving modality-specific characteristics in the multi-modal fusion.
Peng~\etal~\cite{amcmoe} introduce a novel top-down dynamic fusion mechanism that adaptively integrates multi-modal information, addressing the variation in data quality across different modalities and patients in various medical domains.
Wang~\etal~\cite{wang2024demo} propose the DeMo, which first introduces the MoE framework to the multi-modal ReID task by combining modality-specific experts to decouple shared and specific fusion features. 
Despite this, these methods still lack fine-grained semantic understanding. 
To address this limitation, we propose text-modulated semantic experts and context-shared structure experts to jointly capture semantic and structural features, enabling more comprehensive identity representation of multi-modalities.

\begin{figure*}
\centering
\includegraphics[width=1.0\linewidth]{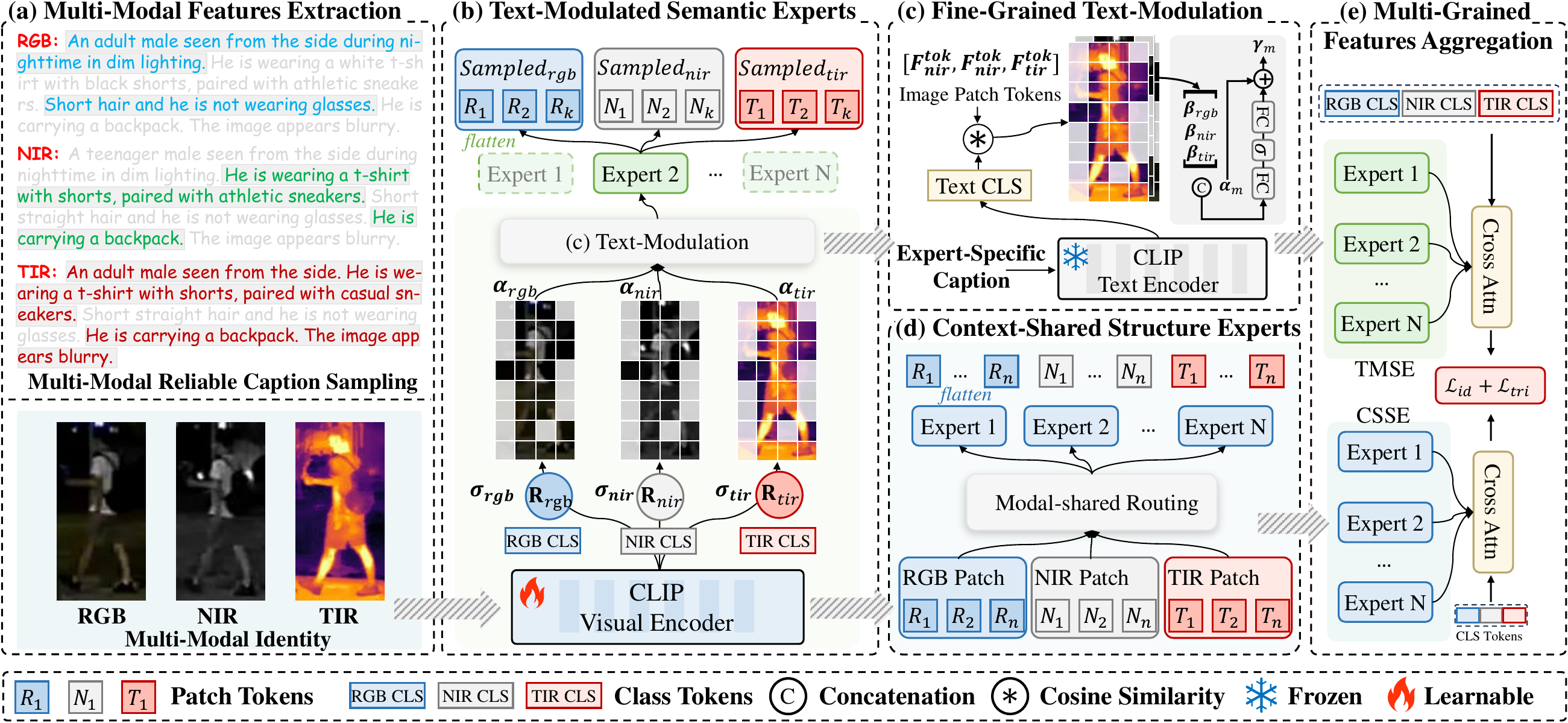}
\caption{
The overview of our proposed framework. 
First, we divide multi-modal features into (a) reliable caption sampling and multi-modal identity representations. 
Then, we feed visual modalities into the CLIP visual encoder to obtain visual embeddings, which are passed into both (b) the Text-Modulated Semantic Experts (TMSE) and (d) Context-Shared Structure Experts (CSSE), decoupling identity recognition into fine-grained appearance recognition and coarse-grained structure recognition. 
For TMSE, we input randomly sampled captions into (c) the Fine-Grained Text-Modulation to guide the sampling process. 
Finally, (e) the Multi-Grained Features Aggregation (MGFA) efficiently integrates multi-grained expert features to form the final identity representation.
}
\label{fig:framework}
\end{figure*}

\section{Methodology}
\label{sec:method}
In this section, we elaborate on the proposed framework, NEXT. 
As shown in Fig.~\ref{fig:framework}, we propose the Multi-Grained Mixture-of-Experts framework that includes the Text-Modulated Semantic Experts (TMSE), the Context-Shared Structure Experts (CSSE), and the Multi-Grained Features Aggregation (MGFA) module.

\subsection{Multi-Modal Features Extraction}
Each multi-modal sample includes three modalities: Visible Light (RGB), Near Infrared (NIR), and Thermal Infrared (TIR), denoted as $\textbf{X}_{I}$ $=$ $[X_{I,rgb},X_{I,nir},X_{I,tir}]$, along with detailed appearance captions $\textbf{X}_{T}$ $=$ $[X_{T,rgb},X_{T,nir},X_{T,tir}]$ which generated by our pipleine in Sec.\ref{sec:caption}.
We feed the visual modality $\textbf{X}_{I}$ into the CLIP visual encoder $\mathcal{V}(\cdot)$ to extract visual features $\mathbf{F}_{I}$ $=$ $[\mathbf{f}_{I}^{\mathrm{cls}}; \mathbf{F}_{I}^{\mathrm{tok}}] \in \mathbb{R}^{3 \times (1+N) \times D}$, and input the appearance description $\textbf{X}_{T}$ into the frozen CLIP text encoder $\mathcal{T}(\cdot)$ to extract textual features $\mathbf{F}_{T}$ $=$ $[\mathbf{f}_{T}^{\mathrm{cls}}; \mathbf{F}_{T}^{\mathrm{tok}}] \in \mathbb{R}^{3 \times (1+L) \times D}$, where $N$ denotes the number of visual patch tokens, $L$ is the length of text tokens, $D$ is the dimension length.

\subsection{Text-Modulated Semantic Experts}
In real world, challenges like illumination variations, background noise, and image degradation impact quality of local regions across modalities, making it essential to bridge modality gaps and extract complementary features.
However, existing approaches often lack explicit semantic guidance. 
To address this, we leverage the object-level textual semantics $\mathbf{X}_T$ extracted from MLLMs to guide a set of semantic sampling experts $\textbf{E}_{T} = \{ E_{t}^{(1)}, E_{t}^{(2)}, \dots, E_{t}^{(N_T)}\}$ in dynamically select semantic-related local features across different modalities, to tackle various semantic challenges, where $N_T$ denotes the number of semantic experts. 
The structure of each expert is defined as:
\begin{equation}
E_{t}^{(i)}(\mathbf{F}) = \mathrm{Dropout}\left( \mathrm{MLP}\left( \mathrm{LN}(\mathbf{F}) \right) \right) + \mathbf{F},
\label{eq:expert}
\end{equation}
where $\mathbf{F}$ denotes the input features, $\mathrm{LN}$ denotes a layer normalization, $\mathrm{MLP}$ refers a multilayer perceptron, and $\mathrm{Dropout}$ with a rate of 0.1 is applied to prevent overfitting.

\textbf{Dynamic Sampling Routing.}
To sample semantic features from each modality, we design a set of modality-specific sampling routes for each expert, denoted as $\mathbf{R}=[\mathbf{R}_{rgb}, \mathbf{R}_{nir}, \mathbf{R}_{tir}]$. 
The patch-level features $\mathbf{F}_{I,m}^{\mathrm{tok}}$ are encoded into the routing matrix $\boldsymbol{\alpha}_m$, and the class token $\mathbf{f}_{I,m}^{\mathrm{cls}}$ are mapped into the dynamic threshold $\boldsymbol{\sigma}_m$ as follows:
\begin{equation}
\begin{aligned}
    \boldsymbol{\alpha}_{m} &= \mathrm{FC^2}\left( \delta\left( \mathrm{FC^1}(\mathbf{F}_{I,m}^{\mathrm{tok}}) \right) \right), \quad \boldsymbol{\alpha}_{m} \in \mathbb{R}^{H \times W}\\
    \boldsymbol{\sigma}_{m} &= \mathrm{FC^2}\left( \delta\left( \mathrm{FC^1}(\mathbf{f}_{I,m}^{\mathrm{cls}}) \right) \right), \quad 
    \boldsymbol{\sigma}_{m} \in \mathbb{R}
\end{aligned}
\label{eq:sampling_router}
\end{equation}
where $\mathrm{FC^1 \in \mathbb{R}^{D \times D}}$ and $\mathrm{FC^2 \in \mathbb{R}^{D \times 1}}$ denotes fully connected layer, $\delta$ is GELU activation function~\cite{hendrycks2016gelu}, and $m \in \{rgb, nir, tir\}$. 
Then, we calculate the sampling matrix $\mathbf{M}_m$ to capture the key tokens for current expert,
\begin{equation}
\mathbf{M}_m(i, j) = 
\begin{cases}
1, & \text{if } \boldsymbol{\alpha}_m(i, j) > \boldsymbol{\sigma}_{m} \\
0, & \text{otherwise}
\end{cases},
\label{eq:sampling_mask}
\end{equation}
as formulated in Eq.~(\ref{eq:sampling_mask}), we assign $\mathrm{1}$ to entries in $\boldsymbol{\alpha}_m$ that exceed the threshold $\boldsymbol{\sigma}_m$ and $\mathrm{0}$ otherwise, to discard irrelevant tokens while keeping the operation differentiable.

\textbf{Fine-Grained Text Modulation.} 
To empower the semantic awareness of each expert, we employ random semantic text prompts during training to guide each expert focus on distinct semantic challenges. 
Formally, the raw textual caption is decomposed into fine-grained sentences for each modality: $X_{T,m} = \{ s_m^{(1)}, s_m^{(2)}, \dots, s_m^{(n_m)} \}$, where $n_m$ is the maximum number of sentences. 
For each expert $E_t^{(i)}$, we randomly select a subset $\hat{X}_{T,m}^{(i)}$ from $X_{T,m}$ as its modulation signal. 
The sampling process is defined as:
\begin{equation}
\begin{split}
    \hat{X}_{T,m}^{(i)} \sim \mathbf{Sampling}(X_{T,m}), \\
    \hat{\mathbf{X}}_{T}^{(i)} = [\hat{X}_{T,rgb}^{(i)}, \hat{X}_{T,nir}^{(i)}, \hat{X}_{T,tir}^{(i)}],
\end{split}
\end{equation}
where $\mathrm{i}$ denotes the $\mathrm{i}$-th semantic experts, and $\mathbf{Sampling}$ is a random probability distribution which controls the diversity and consistency of modulated signals.

To modulate the sampling route, we first encode the randomly sampled high-quality semantic text into the visual-text latent space as text feature vector $\mathbf{f}^{\mathrm{cls}}_{\hat{T},m}$. 
Second, we compute the cosine similarity between this text vector and the visual patch tokens to obtain the semantic matrix $\boldsymbol{\beta}_{m}$:
\begin{equation}
    \boldsymbol{\beta}_{m} = \frac{\mathbf{f}^{\mathrm{cls}}_{\hat{T},m} \cdot \mathbf{F}^{\mathrm{tok}}_{I,m}}{\|\mathbf{f}^{\mathrm{cls}}_{\hat{T},m}\|_2 \cdot \|\mathbf{F}^{\mathrm{tok}}_{I,m}\|_2}, \quad \boldsymbol{\beta}_{m} \in \mathbb{R}^{H \times W}
\end{equation}
where $\|\cdot\|_2$ means $L_2$ normalization.
Then, the semantic guidance is integrated with the route matrix $\boldsymbol{\alpha}_{m}$ to produce the modulation matrix $\boldsymbol{\gamma}_{m}$:
\begin{equation}
\begin{aligned}
\boldsymbol{\gamma}_{m} &= \mathrm{FC^4}\left( \delta \left( \mathrm{FC^3}(\boldsymbol{\alpha}_{m}, \boldsymbol{\beta}_{m}) \right) \right) + \boldsymbol{\alpha}_{m}, \boldsymbol{\gamma}_{m} \in \mathbb{R}^{H \times W}\\
\end{aligned}
\end{equation}
where $\mathrm{FC^3 \in \mathbb{R}^{2 \times D/2}}$ and $\mathrm{FC^4 \in \mathbb{R}^{D/2 \times 1}}$ denotes the fully connected layer within the modulation network whose structure is illustrated in Fig.~\ref{fig:framework}\textcolor{blue}{(c)}.

Finally, we replace the route matrix $\boldsymbol{\alpha}_{m}$ in Eq.~(\ref{eq:sampling_mask}) with the modulated matrix $\boldsymbol{\gamma}_{m}$ to obtain the final modulation-based sampling matrix $\hat{\mathbf{M}}_{m}$.

\begin{equation}
\begin{aligned}
\hat{\mathbf{F}}_{I,m}^{\mathrm{tok}} 
&= E_{t}(\mathbf{F}_{I,m}^{\mathrm{tok}}), 
\hat{\mathbf{F}}_{I} 
= \mathrm{Concat} \left( 
\hat{\mathbf{M}}_{m} \odot \hat{\mathbf{F}}_{I,m}^{\mathrm{tok}} 
\right),
\end{aligned}
\label{eq:semantic_feature}
\end{equation}
According to Eq.~(\ref{eq:semantic_feature}) above, $\hat{\mathbf{M}}_{m}$ dynamically samples the informative features for each semantic expert $\mathrm{E}^{(i)}_t$ within a modality. 
These sampled features are concatenated to obtain the final multi-modal semantic feature $\hat{\mathbf{F}}_{I}$.

\subsection{Context-Shared Structure Experts}
Although semantic experts focus on fine-grained part sampling under various semantic challenges. 
The identity recognition and discrimination still rely on the structural integrity of the object and the environmental context, such as viewpoint, illumination, and capture time. 
To this end, we introduce the structure aware experts, denoted as $\mathbf{E}_{C} = \{ E_{c}^{(1)}, E_{c}^{(2)}, \dots, E_{c}^{(N_C)}\}$, to comprehensively perceive the holistic object structure across modalities, where $N_C$ denotes the number of structure experts and $E_c^{(i)}$ shares the same architecture as $E_t^{(i)}$ in Eq.~(\ref{eq:expert}).

Specifically, we first concatenate multi-modal patch features as $\mathbf{F}_I^{\mathrm{tok}} = [\mathbf{F}_{I,rgb}^{\mathrm{tok}}, \mathbf{F}_{I,nir}^{\mathrm{tok}}, \mathbf{F}_{I,tir}^{\mathrm{tok}}]$, and then feed them into a modality-shared routing network $\mathbf{R}_s$ to compute the expert fusion weights as follows:
\begin{equation}
\bm{\omega} = \mathbf{Softmax} \left( \mathrm{FC} \left( \mathbf{F}_I^{\text{tok}} \right) \right), \quad \bm{\omega} \in \mathbb{R}^{N_C},
\end{equation}
where $\mathbf{Softmax}$ encodes the expert selection weights, resulting in the soft routing matrix $\bm{\omega}$.

Finally, we feed the concatenated multi-modal token features into the structure experts $\mathrm{E}_{C}$ and compute the final multi-modal structural representation by weighting each output of the expert using the soft routing matrix $\bm{\omega}$. 
The process is formulated as follows:
\begin{equation}
\begin{aligned}
\tilde{\mathbf{F}}_{I} &= \sum_{i=1}^{N_C} \omega_i \cdot E^{(i)}_{c}(\mathbf{F}_I^{\text{tok}}), \\
\tilde{\mathbf{F}}_{I}^{\text{tok}} &= \left\{ E^{(i)}_{c}(\mathbf{F}_I^{\text{tok}}) \mid i = 1, 2, \dots, N_C \right\},
\end{aligned}
\end{equation}
where $\tilde{\mathbf{F}}_{I}$ represents the final multi-modal structure feature.

\subsection{Multi-Grained Features Aggregation}
Different types of experts mine identity features at varying levels of granularity.
The semantic experts $\mathrm{E}_{T}$ extract fine-grained semantic features $\hat{\mathbf{F}}_{I}^{(i)}$, and the structural experts $\mathrm{E}_{C}$ capture coarse-grained structural consistency features $\tilde{\mathbf{F}}_{I}$. 
As illustrated in Fig.~\ref{fig:framework}\textcolor{blue}{(e)}, we combine these features to form the complete object identity feature set $\mathbf{F}_{I}^{\mathrm{exp}}=[\hat{\mathbf{F}}_{I}^{(1)},\hat{\mathbf{F}}_{I}^{(2)},\dots,\hat{\mathbf{F}}_{I}^{(N_T)};\tilde{\mathbf{F}}_{I}]$. 
The modality class tokens are concatenated to form the modality query features $\mathbf{f}_{I}^{\mathrm{cls}} = [ \mathbf{f}_{I,rgb}^{\mathrm{cls}}, \mathbf{f}_{I,nir}^{\mathrm{cls}}, \mathbf{f}_{I,tir}^{\mathrm{cls}}]$.
Following this, we treat $\mathbf{f}_{I}^{\mathrm{cls}}$ as the query features $\mathrm{\textit{Q}}$ and the expert features as the key features $\mathrm{\textit{K}}$ and value features $\mathrm{\textit{V}}$.
Through the Cross-Attention mechanism~\cite{DBLP:conf/iclr/DosovitskiyB0WZ21}, we obtain the multi-modal representation for each expert and average them to form the final identity representation $\hat{\mathbf{f}}_{I}^{\mathrm{cls}}$. 
The process is formulated as follows:
\begin{equation}
\hat{\mathbf{f}}_{I}^{\mathrm{cls}}=\mathrm{Average}\left(\hat{\mathbf{f}}_{I}^{\mathrm{cls(1)}},\hat{\mathbf{f}}_{I}^{\mathrm{cls(2)}},\dots,\hat{\mathbf{f}}_{I}^{\mathrm{cls(N_T)}};\tilde{\mathbf{f}}_{I}^{\mathrm{cls}}\right),
\end{equation}
\begin{equation}
\begin{split}
\hat{\mathbf{f}}_{I}^{\mathrm{cls(i)}} ={}& \mathbf{FFN}\Big(
    \mathrm{LN}\Big(
    \mathbf{CA}\big( 
    \mathbf{f}_{I}^{\mathrm{cls}}, 
    \mathbf{F}_{I}^{\mathrm{exp(i)}}
    \big)
    \Big)
\Big), \\
& \text{for } i = 1, \dots, N_T+1
\end{split}
\end{equation}
where $\mathbf{CA}$ is the Cross-Attention mechanism, and $\mathbf{FFN}$ is the Feed-Forward Network.

\subsection{Optimization and Inference}
In line with prior works~\cite{DBLP:conf/iccv/He0WW0021,DBLP:conf/cvpr/0004GLL019}, we use the object identity ID label as the ground-truth to supervise the classification loss $\mathcal{L}_{id}$, and adopt the triplet loss $\mathcal{L}_{tri}$ to enhance identity distribution compactness and separability. 
The final loss function is defined as:
\begin{equation}
\mathcal{L}_{final}(\hat{\mathbf{f}}_{I}^{\mathrm{cls}}) = \mathcal{L}_{id}(\hat{\mathbf{f}}_{I}^{\mathrm{cls}}) + \mathcal{L}_{tri}(\hat{\mathbf{f}}_{I}^{\mathrm{cls}}).
\end{equation}
During the inference, the fused feature serves as the final object representation for multi-modal retrieval.

\begin{figure}[t]
\centering
\includegraphics[width=1.0\linewidth]{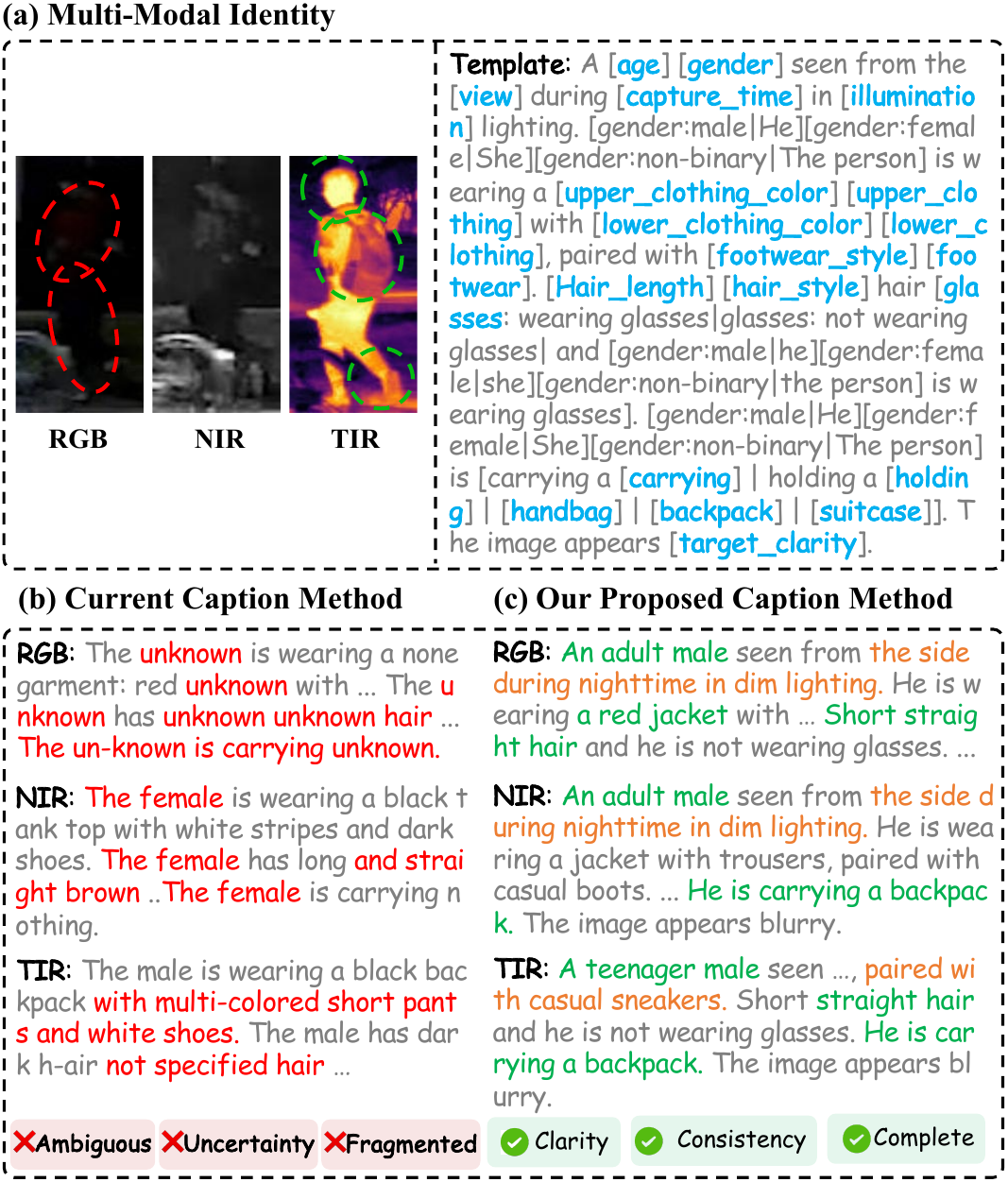}
\caption{
Generated caption samples and text quality comparisons on the person dataset.
}
\label{fig:0_motivation_4}
\end{figure}

\begin{figure*}
\centering
\includegraphics[width=1.0\linewidth]{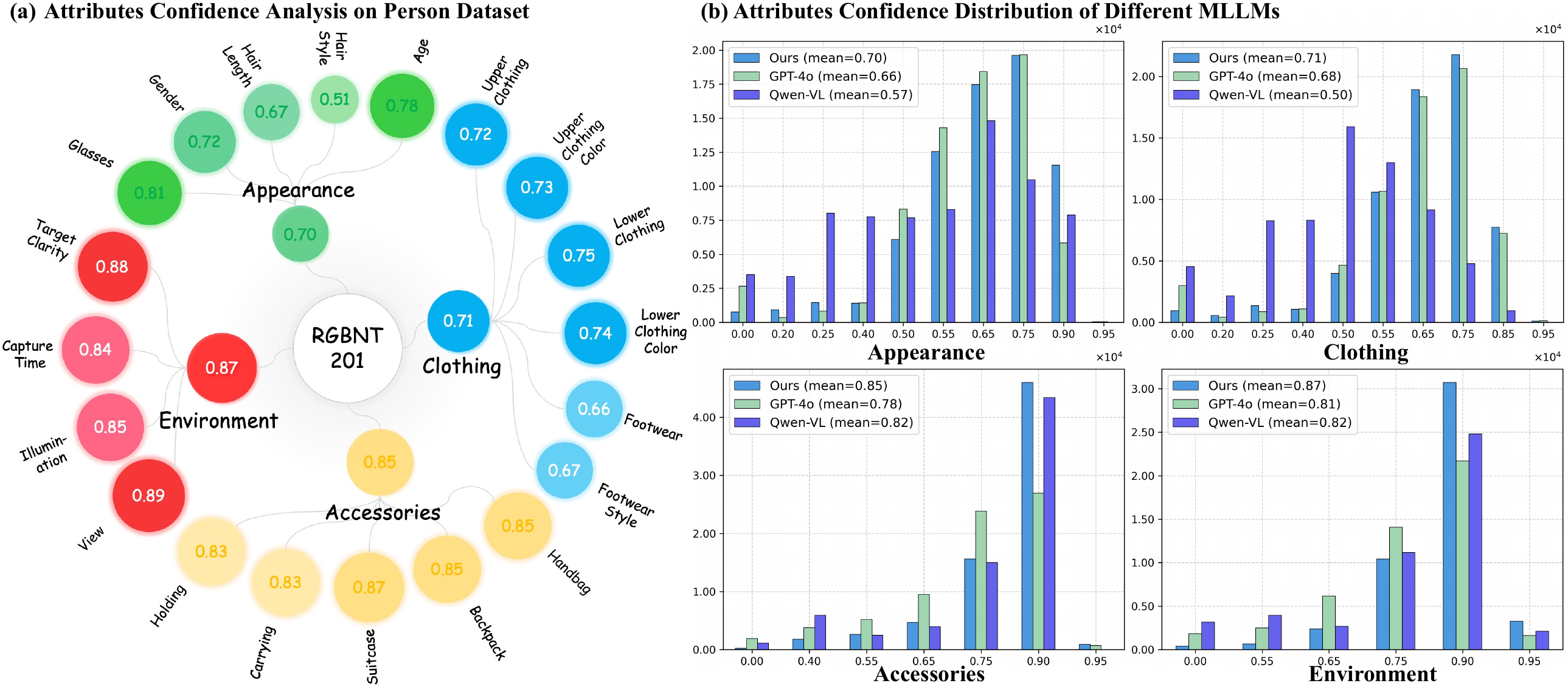}
\caption{
In the person dataset RGBNT201, we divide attributes into four groups: Appearance, Clothing, Accessories, and Environment. 
(a) Most attributes are reliable and have a confidence level higher than 0.7. The size of the circle and the shade of the color represent the level of confidence.
(b) The confidence distribution of our method is significantly higher than the GPT-4o and Qwen-VL native output on four different types of attributes.
Better view with colors and zooming in.
}
\label{fig:0_motivation_3}
\end{figure*}

\section{Reliable Caption Generation}
\label{sec:caption}
Thanks to the powerful semantic understanding capability of Multi-modal Large Language Models (MLLMs), we leverage their out-of-the-box ability to generate comprehensive textual semantic descriptions of the object appearance.
However, simple template-based text generation methods~\cite{wang2025idea,hu2024empowering,DBLP:conf/aaai/ZhaiZH0JC24} face many practical challenges when dealing with multi-modal data, such as noise interference from low-quality images and significant style discrepancies in infrared modalities, making it hard to observe the object's fine-grained details. 
To overcome these challenges, we first employ multiple MLLMs to generate several attributes with confidence scores, and then leverage the confidence scores with a multi-modal attributes aggregation strategy to generate reliable and comprehensive identity captions.

\subsection{Confidence-Aware Attribute Generation}
To flexibly analyze the object appearance, we set the generation of identity attributes as the basic step in caption generation. 
We first define an instruction template to guide multiple MLLMs (\textit{e.g.}, GPT-4o~\cite{hurst2024gpt4o}, Qwen-VL~\cite{bai2023qwenvl}, etc.) in structuring the attribute set of the input image. 
However, simple structured attribute information cannot quantitatively measure the confidence level of each item, the model suffers from low-quality visual modality noise, which leads to incorrect attribute selection. 
To address this, we introduce the concept of \textit{confidence} score in the instruction template.
%
This allows us to observe the confidence level of each MLLM assigns to its output for each attribute. 
Additionally, to avoid the model ignoring environment information, we not only structure the output of `\textcolor{attr_color}{\textit{age}}', `\textcolor{attr_color}{\textit{gender}}', `\textcolor{attr_color}{\textit{upper clothing}}', `\textcolor{attr_color}{\textit{lower clothing}}', `\textcolor{attr_color}{\textit{hairstyle}}', `\textcolor{attr_color}{\textit{footwear}}' and other appearance attributes, but also require the MLLMs to output `\textcolor{attr_color}{\textit{view}}', `\textcolor{attr_color}{\textit{illumination}}', `\textcolor{attr_color}{\textit{capture time}}', and `\textcolor{attr_color}{\textit{target clarity}}' attributes to aware environmental context.

\subsection{Multi-Modal Attribute Aggregation}
Based on the confidence-aware attribute set with confidence score, we are able to further generate reliable captions for the object. 
However, as shown in Fig.~\ref{fig:motivation_2}\textcolor{blue}{(a)} and Fig.~\ref{fig:0_motivation_4}, the RGB modal is affected by lighting degradation, causing the MLLMs to fail to recognize the `\textcolor{attr_color}{\textit{backpack}}' attribute. 
In contrast, the thermal infrared modality for the same identity easily reveals additional `\textcolor{attr_color}{\textit{backpack}}' attribute. 
This leads to the model generating vague and incomplete textual captions such as `\textcolor{attr_color2}{\textit{none}},' `\textcolor{attr_color2}{\textit{unknown}},' or `\textcolor{attr_color2}{\textit{unclear}}' based on monocular visual modality. 
To address this problem, we employ multiple MLLMs to recognize attributes and output confidence scores. As shown in Fig.~\ref{fig:0_motivation_3}, GPT-4o produces higher confidence on fine-grained appearance and clothing attributes, while Qwen-VL focuses more on accessories and environment cues.
Therefore, we propose a multi-modal attribute aggregation strategy based on confidence scores to preprocess complementary semantic information across different visual modalities.

\textbf{Multi-Modal Merge and Complement.} 
Specifically, we first use multiple MLLMs to identify object attributes, and for each identified attribute, we select the MLLMs with the highest confidence as their final value to improve the accuracy of each attribute within the modality.
Then, taking the RGB modality as an example, its modality-specific low-frequency attributes, such as `\textcolor{attr_color}{\textit{illumination}},' `\textcolor{attr_color}{\textit{upper clothing color}},' and `\textcolor{attr_color}{\textit{lower clothing color}},' are easy to recognize. 
In contrast, attributes requiring high-frequency information, such as `\textcolor{attr_color}{\textit{carrying}},' `\textcolor{attr_color}{\textit{holding}},' `\textcolor{attr_color}{\textit{handbag}},' and `\textcolor{attr_color}{\textit{backpack}},' are often incorrectly recognized as `\textcolor{attr_color2}{\textit{unknown}},' `\textcolor{attr_color2}{\textit{unclear}},' or `\textcolor{attr_color2}{\textit{not carrying}}.' 
This significantly affects the clarity of the final generated captions.
To address this, we rely on the confidence scores to select the highest-confidence attributes from the NIR and TIR modalities, which are unaffected by lighting conditions, for rechecking and complementing missing information.
Similarly, we apply the same complementary strategy to each modality in order to obtain a complete and reliable set of attributes for each modality. 

\textbf{Caption Generation.} Building on the above complete attribute set, we define the template-based instruction to guide the LLM~\cite{liu2024deepseek} in combining the confidence-aware attribute set from each modality to generate the final text caption.
Therefore, we have constructed a complete and reliable caption generation pipeline, and vehicle datasets are applied same process to obtain comprehensive semantic description of the vehicle appearance.

\begin{table}[ht]
\renewcommand\arraystretch{1.2}
\centering
\caption{
Statistical analysis of five datasets.
}
\resizebox{1.0\columnwidth}{!}{
\begin{tabular}{c|c|ccc}
\noalign{\hrule height 0.8pt}
\textbf{Datasets} & \textbf{Object Type} & \textbf{\# Samples} & \textbf{\# IDs} & \textbf{\# Cams} \\
\hline
\hline
RGBNT201~\cite{DBLP:conf/aaai/ZhengWCLT21} & Person & 4,787 & 201 & 4 \\
Market-MM~\cite{DBLP:conf/aaai/WangLZHT22} & Person & 32,668 & 1,501 & 6 \\
\hline
MSVR310~\cite{DBLP:journals/inffus/ZhengZMLTM23} & Vehicle &2,087 & 310 & 8 \\
RGBNT100~\cite{DBLP:conf/aaai/Li0ZZ020} & Vehicle & 17,250 & 100 & 8 \\
WMVEID863~\cite{DBLP:journals/inffus/ZhengMSWLT25} & Vehicle & 4,709 & 863 & 8 \\
\noalign{\hrule height 0.8pt}
\end{tabular}
}
\label{tb:datasets}
\end{table}

\begin{table*}[t]
\centering
\renewcommand\arraystretch{1.2}
\caption{
Comparison with the state-of-the-art methods on person datasets: RGBNT201, Market-MM about mAP(\%) and Rank-K(\%). 
The best and second best results are marked in \textbf{bold} and \underline{underline}, respectively.
}
\resizebox{2.0\columnwidth}{!}{
\begin{tabular}{c|l|c|c||cccc|cccc}
\noalign{\hrule height 0.8pt}
&
\multirow{2}{*}{\textbf{Methods}} & 
\multirow{2}{*}{\textbf{Venue}} & 
\multirow{2}{*}{\textbf{Backbone}} & 
\multicolumn{4}{c|}{\textbf{RGBNT201}} & 
\multicolumn{4}{c}{\textbf{Market-MM}} \\
\cline{5-12}
& & & & \textbf{mAP} & \textbf{R-1} & \textbf{R-5} & \textbf{R-10} 
& \textbf{mAP} & \textbf{R-1} & \textbf{R-5} & \textbf{R-10} \\
\hline
\hline
\multirow{5}{*}{\rotatebox{90}{\textbf{Single-Modal}}}
& HACNN~\cite{DBLP:conf/cvpr/LiZG18} & CVPR'18 & ResNet & 21.3 & 19.0 & 34.1 & 42.8 & 42.9 & 69.1 & 86.6 & 92.2 \\
&  \cellcolor{table_color2}MLFN~\cite{DBLP:conf/cvpr/ChangHX18} & \cellcolor{table_color2}CVPR'18 & \cellcolor{table_color2}ResNet & \cellcolor{table_color2}26.1 & \cellcolor{table_color2}24.2 & \cellcolor{table_color2}35.9 & \cellcolor{table_color2}44.1 & \cellcolor{table_color2}42.7 & \cellcolor{table_color2}68.1 & \cellcolor{table_color2}87.1 & \cellcolor{table_color2}92.0 \\
& OSNet~\cite{DBLP:conf/iccv/ZhouYCX19} & ICCV'19 & ResNet & 25.4 & 22.3 & 35.1 & 44.7 & 39.7 & 69.3 & 86.7 & 91.3 \\
& \cellcolor{table_color2}CAL~\cite{DBLP:conf/iccv/Rao0L021} & \cellcolor{table_color2}ICCV'21 & \cellcolor{table_color2}ResNet & \cellcolor{table_color2}27.6 & \cellcolor{table_color2}24.3 & \cellcolor{table_color2}36.5 & \cellcolor{table_color2}45.7 & \cellcolor{table_color2}- & \cellcolor{table_color2}- & \cellcolor{table_color2}- & \cellcolor{table_color2}- \\
& TransReID~\cite{DBLP:conf/iccv/He0WW0021} & ICCV'21 & ViT-B/16 & 63.8 & 65.8 & 78.5 & 83.9 & 73.0 & 88.9 & 95.8 & 97.6 \\
\hline
\multirow{12}{*}{\rotatebox{90}{\textbf{Multi-Modal}}}
& \cellcolor{table_color2}HAMNet~\cite{DBLP:conf/aaai/Li0ZZ020} & \cellcolor{table_color2}AAAI'20 & \cellcolor{table_color2}ResNet & \cellcolor{table_color2}27.7 & \cellcolor{table_color2}26.3 & \cellcolor{table_color2}41.5 & \cellcolor{table_color2}51.7 & \cellcolor{table_color2}60.0 & \cellcolor{table_color2}82.8 & \cellcolor{table_color2}92.5 & \cellcolor{table_color2}95.0 \\
& PFNet~\cite{DBLP:conf/aaai/ZhengWCLT21} & AAAI'21 & ResNet & 38.5 & 38.9 & 52.0 & 58.4 & 60.9 & 83.6 & 92.8 & 95.5 \\
& \cellcolor{table_color2}IEEE~\cite{DBLP:conf/aaai/WangLZHT22} & \cellcolor{table_color2}AAAI'22 & \cellcolor{table_color2}ResNet & \cellcolor{table_color2}46.4 & \cellcolor{table_color2}47.1 & \cellcolor{table_color2}58.5 & \cellcolor{table_color2}64.2 & \cellcolor{table_color2}64.3 & \cellcolor{table_color2}83.9 & \cellcolor{table_color2}93.0 & \cellcolor{table_color2}95.7 \\
& HTT~\cite{DBLP:conf/aaai/WangHZ024} & AAAI'24 & ViT-B/16 & 71.1 & 73.4 & 83.1 & 87.3 & 67.2 & 81.5 & 95.8 & 97.8 \\
& \cellcolor{table_color2}TOP-ReID~\cite{DBLP:conf/aaai/WangLZLTL24} & \cellcolor{table_color2}AAAI'24 & \cellcolor{table_color2}ViT-B/16 & \cellcolor{table_color2}72.3 & \cellcolor{table_color2}76.6 & \cellcolor{table_color2}84.7 & \cellcolor{table_color2}89.4 & \cellcolor{table_color2}82.0 & \cellcolor{table_color2}92.4 & \cellcolor{table_color2}97.6 & \cellcolor{table_color2}98.6 \\
& EDITOR~\cite{zhang2024magic} & CVPR'24 & ViT-B/16 & 66.5 & 68.3 & 81.1 & 88.2 & 77.4 & 90.8 & 96.8 & 98.3 \\
& \cellcolor{table_color2}ICPL-ReID~\cite{li2025icpl} & \cellcolor{table_color2}T-MM'25 & \cellcolor{table_color2}CLIP-B/16 & \cellcolor{table_color2}75.1 & \cellcolor{table_color2}77.4 & \cellcolor{table_color2}84.2 & \cellcolor{table_color2}87.9 & \cellcolor{table_color2}\underline{85.1} & \cellcolor{table_color2}\underline{94.7} & \cellcolor{table_color2}\underline{98.4} & \cellcolor{table_color2}\underline{99.1} \\
& MambaPro~\cite{Wang2024MambaPro} & AAAI'25 & CLIP-B/16 & 78.9 & \underline{83.4} & 89.8 & 91.9 & 84.1 & 92.8 & 97.7 & 98.7 \\
& \cellcolor{table_color2}PromptMA~\cite{10955143} & \cellcolor{table_color2}T-IP'25 & \cellcolor{table_color2}CLIP-B/16 & \cellcolor{table_color2}78.4 & \cellcolor{table_color2}80.9 & \cellcolor{table_color2}87.0 & \cellcolor{table_color2}88.9 & \cellcolor{table_color2}- & \cellcolor{table_color2}- & \cellcolor{table_color2}- & \cellcolor{table_color2}- \\
& DeMo~\cite{wang2024demo} & AAAI'25 & CLIP-B/16 & 79.0 & 82.3 & 88.8 & 92.0 & 83.6 & 93.1 & 97.5 & 98.7 \\
& \cellcolor{table_color2}IDEA~\cite{wang2025idea} & \cellcolor{table_color2}CVPR'25 & \cellcolor{table_color2}CLIP-B/16 & \cellcolor{table_color2}\underline{80.2} & \cellcolor{table_color2}82.1 & \cellcolor{table_color2}\underline{90.0} & \cellcolor{table_color2}\underline{93.3} & \cellcolor{table_color2}- & \cellcolor{table_color2}- & \cellcolor{table_color2}- & \cellcolor{table_color2}- \\
\hline
& \textbf{NEXT} & \textbf{Ours} & \textbf{CLIP-B/16} & \textbf{82.4} & \textbf{86.6} & \textbf{92.0} & \textbf{94.7} & \textbf{85.8} & \textbf{95.3} & \textbf{98.7} & \textbf{99.4} \\
& \cellcolor{table_color}\textit{Improvement} & \cellcolor{table_color}\textcolor{gray}{-} & \cellcolor{table_color}\textcolor{gray}{-} & \cellcolor{table_color}\textcolor{improve_color}{+$\uparrow$2.2} & \cellcolor{table_color}\textcolor{improve_color}{+$\uparrow$3.2} & \cellcolor{table_color}\textcolor{improve_color}{+$\uparrow$2.0} & \cellcolor{table_color}\textcolor{improve_color}{+$\uparrow$1.4} & \cellcolor{table_color}\textcolor{improve_color}{+$\uparrow$0.7} & \cellcolor{table_color}\textcolor{improve_color}{+$\uparrow$0.6} & \cellcolor{table_color}\textcolor{improve_color}{+$\uparrow$0.3} & \cellcolor{table_color}\textcolor{improve_color}{+$\uparrow$0.3} \\
\noalign{\hrule height 0.8pt}
\end{tabular}
}
\label{tb:sota_person}
\end{table*}

\begin{table*}[t]
\centering
\renewcommand\arraystretch{1.2}
\caption{Comparison with the state-of-the-art methods on vehicle datasets: MSVR310, RGBNT100, and WMVEID863 about mAP(\%) and Rank-K(\%). The best and second best results are marked in \textbf{bold} and \underline{underline}, respectively.}
\resizebox{2.0\columnwidth}{!}{
\begin{tabular}{c|l|c|c||cc|cc|cccc}
\noalign{\hrule height 0.8pt}
& 
\multirow{2}{*}{\textbf{Methods}} & 
\multirow{2}{*}{\textbf{Venue}} & 
\multirow{2}{*}{\textbf{Backbone}} & 
\multicolumn{2}{c|}{\textbf{MSVR310}} & 
\multicolumn{2}{c|}{\textbf{RGBNT100}} & 
\multicolumn{4}{c}{\textbf{WMVEID863}} \\
\cline{5-12}
& & & 
& \textbf{mAP} & \textbf{R-1} 
& \textbf{mAP} & \textbf{R-1} 
& \textbf{mAP} & \textbf{R-1} & \textbf{R-5} & \textbf{R-10} \\
\hline
\hline
\multirow{5}{*}{\rotatebox{90}{\textbf{Single-Modal}}}
& \cellcolor{table_color2}BoT~\cite{DBLP:conf/cvpr/0004GLL019} & \cellcolor{table_color2}CVPRW'19 & \cellcolor{table_color2}ResNet & \cellcolor{table_color2}23.5 & \cellcolor{table_color2}38.4 & \cellcolor{table_color2}78.0 & \cellcolor{table_color2}95.1 & \cellcolor{table_color2}51.1 & \cellcolor{table_color2}55.7 & \cellcolor{table_color2}69.8 & \cellcolor{table_color2}74.7 \\
& OSNet~\cite{DBLP:conf/iccv/ZhouYCX19} & ICCV'19 & ResNet & 28.7 & 44.8 & 75.0 & 95.6 & 42.9 & 46.8 & 61.9 & 69.4 \\
& \cellcolor{table_color2}AGW~\cite{ye2021deep} & \cellcolor{table_color2}T-PAMI'21 & \cellcolor{table_color2}ResNet & \cellcolor{table_color2}8.9 & \cellcolor{table_color2}46.9 & \cellcolor{table_color2}73.1 & \cellcolor{table_color2}92.7 & \cellcolor{table_color2}30.3 & \cellcolor{table_color2}35.3 & \cellcolor{table_color2}43.3 & \cellcolor{table_color2}46.5 \\
& PFD~\cite{DBLP:conf/aaai/WangLS0S22} & AAAI'22 & ResNet & 23.0 & 39.9 & 67.5 & 92.6 & 50.2 & 55.3 & 69.8 & 75.3 \\
& \cellcolor{table_color2}TransReID~\cite{DBLP:conf/iccv/He0WW0021} & \cellcolor{table_color2}ICCV'21 & \cellcolor{table_color2}ViT-B/16 & \cellcolor{table_color2}33.4 & \cellcolor{table_color2}48.9 & \cellcolor{table_color2}75.6 & \cellcolor{table_color2}92.9 & \cellcolor{table_color2}67.0 & \cellcolor{table_color2}74.7 & \cellcolor{table_color2}79.5 & \cellcolor{table_color2}82.4 \\
\hline
\multirow{13}{*}{\rotatebox{90}{\textbf{Multi-Modal}}}
& HAMNet~\cite{DBLP:conf/aaai/Li0ZZ020} & AAAI'20 & ResNet & 27.1 & 42.3 & 74.5 & 93.3 & 45.6 & 48.5 & 63.1 & 68.8 \\
& \cellcolor{table_color2}PFNet~\cite{DBLP:conf/aaai/ZhengWCLT21} & \cellcolor{table_color2}AAAI'21 & \cellcolor{table_color2}ResNet & \cellcolor{table_color2}23.5 & \cellcolor{table_color2}37.4 & \cellcolor{table_color2}68.1 & \cellcolor{table_color2}94.1 & \cellcolor{table_color2}50.1 & \cellcolor{table_color2}55.9 & \cellcolor{table_color2}68.7 & \cellcolor{table_color2}75.1 \\
& IEEE~\cite{DBLP:conf/aaai/WangLZHT22} & AAAI'22 & ResNet & 21.0 & 41.0 & 61.3 & 87.8 & 45.9 & 48.6 & 64.3 & 67.9 \\
& \cellcolor{table_color2}CCNet~\cite{DBLP:journals/inffus/ZhengZMLTM23} & \cellcolor{table_color2}INFFUS'23 & \cellcolor{table_color2}ResNet & \cellcolor{table_color2}36.4 & \cellcolor{table_color2}55.2 & \cellcolor{table_color2}77.2 & \cellcolor{table_color2}96.3 & \cellcolor{table_color2}50.3 & \cellcolor{table_color2}52.7 & \cellcolor{table_color2}69.6 & \cellcolor{table_color2}75.1 \\
& TOP-ReID~\cite{DBLP:conf/aaai/WangLZLTL24} &AAAI'24 & ViT-B/16 & 35.9 & 44.6 & 81.2 & 96.4 & 67.7 & 75.3 & 80.8 & 83.5 \\
& \cellcolor{table_color2}FACENet~\cite{DBLP:journals/inffus/ZhengMSWLT25} & \cellcolor{table_color2}INFFUS'25 & \cellcolor{table_color2}ViT-B/16 & \cellcolor{table_color2}36.2 & \cellcolor{table_color2}54.1 & \cellcolor{table_color2}81.5 & \cellcolor{table_color2}96.9 & \cellcolor{table_color2}\underline{69.8} & \cellcolor{table_color2}77.0 & \cellcolor{table_color2}81.0 & \cellcolor{table_color2}84.2 \\
& EDITOR~\cite{zhang2024magic} & CVPR'24 & ViT-B/16 & 39.0 & 49.3 & 82.1 & 96.4 & 65.6 & 73.8 & 80.0 & 82.3 \\
& \cellcolor{table_color2}ICPL-ReID~\cite{li2025icpl} & \cellcolor{table_color2}T-MM'25 & \cellcolor{table_color2}CLIP-B/16 & \cellcolor{table_color2}\underline{56.9} & \cellcolor{table_color2}\underline{77.7} & \cellcolor{table_color2}87.0 & \cellcolor{table_color2}\textbf{98.6} & \cellcolor{table_color2}67.2 & \cellcolor{table_color2}74.0 & \cellcolor{table_color2}81.3 & \cellcolor{table_color2}\underline{85.6} \\
& MambaPro~\cite{Wang2024MambaPro} & AAAI'25 & CLIP-B/16 & 47.0 & 56.5 & 83.9 & 94.7 & 69.5 & 76.9 & 80.6 & 83.8 \\
& \cellcolor{table_color2}PromptMA~\cite{10955143} & \cellcolor{table_color2}T-IP'25 & \cellcolor{table_color2}CLIP-B/16 & \cellcolor{table_color2}55.2 & \cellcolor{table_color2}64.5 & \cellcolor{table_color2}85.3 & \cellcolor{table_color2}97.4 & \cellcolor{table_color2}- & \cellcolor{table_color2}- & \cellcolor{table_color2}- & \cellcolor{table_color2}- \\
& DeMo~\cite{wang2024demo} & AAAI'25 & CLIP-B/16 & 49.2 & 59.8 & 86.2 & 97.6 & 68.8 & \underline{77.2} & \underline{81.5} & 83.8 \\
& \cellcolor{table_color2}IDEA~\cite{wang2025idea}  & \cellcolor{table_color2}CVPR'25 & \cellcolor{table_color2}CLIP-B/16 & \cellcolor{table_color2}47.0 & \cellcolor{table_color2}62.4 & \cellcolor{table_color2}\underline{87.2} & \cellcolor{table_color2}96.5 & \cellcolor{table_color2}- & \cellcolor{table_color2}- & \cellcolor{table_color2}- & \cellcolor{table_color2}- \\
\hline
& \textbf{NEXT} & \textbf{Ours} & \textbf{CLIP-B/16} & \textbf{60.8} & \textbf{79.0} & \textbf{88.2} & \underline{97.7} & \textbf{70.9} & \textbf{77.8} & \textbf{84.3} & \textbf{86.7} \\
& \cellcolor{table_color}\textit{Improvement} & \cellcolor{table_color}\textcolor{gray}{-} & \cellcolor{table_color}\textcolor{gray}{-} & \cellcolor{table_color}\textcolor{improve_color}{+$\uparrow$3.9} & \cellcolor{table_color}\textcolor{improve_color}{+$\uparrow$1.3} & \cellcolor{table_color}\textcolor{improve_color}{+$\uparrow$1.0} & \cellcolor{table_color}\textcolor{gray}{-} & \cellcolor{table_color}\textcolor{improve_color}{+$\uparrow$1.1} & \cellcolor{table_color}\textcolor{improve_color}{+$\uparrow$0.6} & \cellcolor{table_color}\textcolor{improve_color}{+$\uparrow$2.8} & \cellcolor{table_color}\textcolor{improve_color}{+$\uparrow$1.1} \\
\noalign{\hrule height 0.8pt}
\end{tabular}
}
\label{tb:sota_vehicle}
\end{table*}

\section{Experiment}
\subsection{Datasets and Evaluation Protocols}
\noindent

\subsubsection{Datasets}
As shown in Table~\ref{tb:datasets}, we evaluate our method on five multi-modal object ReID datasets: RGBNT201~\cite{DBLP:conf/aaai/ZhengWCLT21}, Market-MM~\cite{DBLP:conf/aaai/WangLZHT22}, MSVR310~\cite{DBLP:journals/inffus/ZhengZMLTM23}, RGBNT100~\cite{DBLP:conf/aaai/Li0ZZ020}, and WMVEID863~\cite{DBLP:journals/inffus/ZhengMSWLT25}. 
To extend these datasets, we employ GPT-4o~\cite{hurst2024gpt4o} and Qwen-VL~\cite{bai2023qwenvl} to automatically generate object attribute with confidence, and DeepSeek-V3~\cite{liu2024deepseek} to compose the final caption for each image modality.

\textbf{Person datasets.} RGBNT201 comprises 14,361 person images corresponding to 4,787 multi-modal samples of 201 identities, with 141 identities used for training, 30 for validation, and 30 for testing across four distinct viewpoints. 
Market-MM is a synthetic multi-modal person dataset derived from Market1501~\cite{market1501}, containing 32,668 multi-modal samples of 1,501 identities, of which 751 identities with 12,936 triples form the training set and the remaining 750 identities constitute the gallery and query partitions. 

\textbf{Vehicle datasets.}
MSVR310 contains 6,261 images forming 2,087 multi-modal samples of 310 vehicle identities, with 155 identities and 1,032 samples used for training and the remaining 155 identities contributing 1,055 gallery samples and 591 query samples. 
RGBNT100 includes 100 vehicle identities and augments them with 17,250 thermal images, yielding 8,675 multi-modal triples for training and 8,575 triples for testing, from which 1,715 samples are selected as queries. 
WMVeID863 provides 4,709 image triplets of 863 identities captured from eight views, where 603 identities with 3,482 triplets are used for training and 260 identities with 1,226 triplets form the gallery and query sets, with 959 query samples selected. 

\subsubsection{Evaluation Protocols}
Following the convention of community~\cite{DBLP:conf/iccv/He0WW0021,ye2021deep}, we use the Mean Average Precision (mAP) and Rank-K (K = 1, 5, 10) matching accuracy as evaluation metrics. 
As in previous works~\cite{DBLP:conf/aaai/ZhengWCLT21,DBLP:conf/aaai/Li0ZZ020,DBLP:journals/inffus/ZhengMSWLT25}, we adopt the common evaluation protocol for RGBNT201, Market-MM, RGBNT100 and WMVEID863. 
For MSVR310, we filter out samples with the same identity and time span based on time labels to avoid easy matching~\cite{DBLP:journals/inffus/ZhengZMLTM23}.

\subsection{Implementation Details}
We resize each spectral image to \(256\times128\) for person samples, and \(128\times256\) for vehicle samples. 
Data augmentation includes random horizontal flipping, padding, cropping, and erasing~\cite{DBLP:conf/aaai/Zhong0KL020}. 
All parameters in the CLIP text encoder~\cite{DBLP:conf/icml/RadfordKHRGASAM21} are frozen for text semantic projection, while the CLIP visual branch is fully trainable. 
Adam~\cite{kingma2014adam} optimizer is used with the learning rate of 3.5\textit{e}-4, weight decay of 0.0001, and momentum set to 0.9. 
All experiments are conducted on one NVIDIA RTX 4090 GPU using the PyTorch toolkits.

\subsection{Comparison with State-of-the-Art Methods}
\textbf{Performance on Person Dataset.} 
In Table ~\ref{tb:sota_person}, we compare our NEXT with existing methods on the RGBNT201 and Market-MM datasets. 
Benefiting from the flexible fusion of semantic and structural experts, our method achieves a significant performance lead, reaching 82.4\%/86.6\% mAP and Rank-1 accuracy. 
Compared with DeMo~\cite{wang2024demo}, which adopts the MoE structure to decouple modality-shared and specific features, our method demonstrates superior performance.
Against IDEA~\cite{wang2025idea}, which utilizes semantic inversion and deformable offset sampling, our method leverages flexible text-modulation to guide expert sampling multi-modal semantics, improving mAP and Rank-1 by +2.2\% and +4.5\%, respectively.
On the Market-MM dataset, NEXT also establishes clear advantages, reaching 85.8\%/95.3\% mAP and Rank-1 accuracy. 
%
%
The consistent improvements on the two datasets demonstrate the robustness of NEXT in learning discriminative identity features under diverse multi-modal conditions.

\begin{table*}[!ht]
\centering
\caption{Ablation studies of different modules on person dataset RGBNT201 and vehicle dataset MSVR310.}
\renewcommand\arraystretch{1.2}
\resizebox{1.9\columnwidth}{!}{
\begin{tabular}{c|ccc|cccc|cccc|cc}
\noalign{\hrule height 0.8pt}
\multirow{2}{*}{\textbf{Index}} & \multicolumn{3}{c|}{\textbf{Modules}} & \multicolumn{4}{c|}{\textbf{RGBNT201}}& \multicolumn{4}{c|}{\textbf{MSVR310}} & \multicolumn{1}{c}{\textbf{Params}} & \textbf{Flops}\\
\cline{2-14}
& \textbf{MGFA} & \textbf{TMSE} & \textbf{CSSE} & \textbf{mAP} & \textbf{R-1} & \textbf{R-5} & \textbf{R-10} & \textbf{mAP} & \textbf{R-1} & \textbf{R-5} & \textbf{R-10} & \textbf{M} & \textbf{G} \\
\hline
\hline
(a) & \textcolor{gray}{\ding{53}} & \textcolor{gray}{\ding{53}} & \textcolor{gray}{\ding{53}} & 71.0 & 73.6 & 84.2 & 88.2 & 50.1 & 68.9 & 83.4 & 87.8 & 88.3 & 33.3 \\
(b) & \ding{51} & \textcolor{gray}{\ding{53}} & \textcolor{gray}{\ding{53}} & 74.2 & 76.6 & 87.8 & 91.3 & 55.0 & 73.3 & 86.8 & 91.7 & 88.5 & 33.5 \\
(c) & \ding{51} & \ding{51} & \textcolor{gray}{\ding{53}} & 78.9 & 84.4 & 91.1 & 93.2 & 59.7 & 75.1 & 88.2 & 92.0 & 92.5 & 43.7 \\
\rowcolor{table_color}
(d) & \ding{51} & \ding{51} & \ding{51} & \textbf{82.4} & \textbf{86.6} & \textbf{92.0} & \textbf{94.7} & \textbf{60.8} & \textbf{79.0} & \textbf{89.2} & \textbf{92.2} & 94.8 & 44.8 \\
\noalign{\hrule height 0.8pt}
\end{tabular}
}
\label{tab:ablation_310_201}
\vspace{-0.5em}
\end{table*}

\textbf{Performance on Vehicle Dataset.}
Table~\ref{tb:sota_vehicle} outlines the performance of our method on vehicle datasets. 
On the MSVR310 dataset, our approach significantly outperforms existing methods, achieving 60.8\%/79.0\% mAP and Rank-1 accuracy.
This improvement demonstrates that the semantic sampling and structure aware experts can effectively mitigate viewpoint variations.
On the RGBNT100 dataset, our method continues to lead in mAP performance.
For the challenging WMVEID863 dataset, where lighting variations degrade recognition performance, our method still improves over FACENet~\cite{DBLP:journals/inffus/ZhengMSWLT25} by +1.1\%/+0.8\% mAP and Rank-1.
These results confirm the strong generalization and robustness of NEXT in vehicle ReID under complex conditions. 

\subsection{Ablation Studies}
On the RGBNT201 and MSVR310 datasets, we start from a baseline which is built on a three-branch CLIP visual encoder~\cite{DBLP:conf/icml/RadfordKHRGASAM21}, and incrementally incorporate our components to evaluate their specific contributions to the model’s overall performance.
The three main components are: the Multi-Grained Features Aggregation (MGFA), the Text-Modulated Semantic Experts (TMSE), and the Context-Shared Structure Experts (CSSE).

\textbf{Effectiveness of Key Modules.}
As shown in Table~\ref{tab:ablation_310_201}, we conduct ablation studies on both the RGBNT201 and MSVR310 datasets to evaluate the effectiveness of each proposed component.
In Table~\ref{tab:ablation_310_201}\textcolor{blue}{(a)}, the baseline with a three-branch CLIP visual encoder achieves 71.0\%/73.6\% mAP and Rank-1 on RGBNT201 and 50.1\%/68.9\% on MSVR310.
By introducing the MGFA in Table~\ref{tab:ablation_310_201}\textcolor{blue}{(b)}, which fuses the modality-specific features as a unified expert, the performance improves to 74.2\%/76.6\% and 55.0\%/73.3\%, indicating that adaptive feature fusion benefits representation learning.
In Table~\ref{tab:ablation_310_201}\textcolor{blue}{(c)}, adding the TMSE further enhances performance to 78.9\% mAP and 84.4\% Rank-1 on RGBNT201, and brings an additional +4.7\% mAP and +1.8\% Rank-1 improvement on MSVR310, demonstrating the advantage of semantic modulation in sampling fine-grained features.
Finally, Table~\ref{tab:ablation_310_201}\textcolor{blue}{(d)} combines the model with the CSSE and achieves the best results, 82.4\%/86.6\% mAP and Rank-1 on RGBNT201, and 60.8\%/79.0\% on MSVR310, validating that maintaining structural integrity and environment context awareness is beneficial for obtaining discriminative identity features.
These results validate the complementary contributions and generalizability of the MGFA, TMSE, and CSSE modules, confirming their effectiveness in enhancing multi-modal fusion and discriminative representation learning across different datasets.

\begin{figure}
    \centering
    \includegraphics[width=1.0\linewidth]{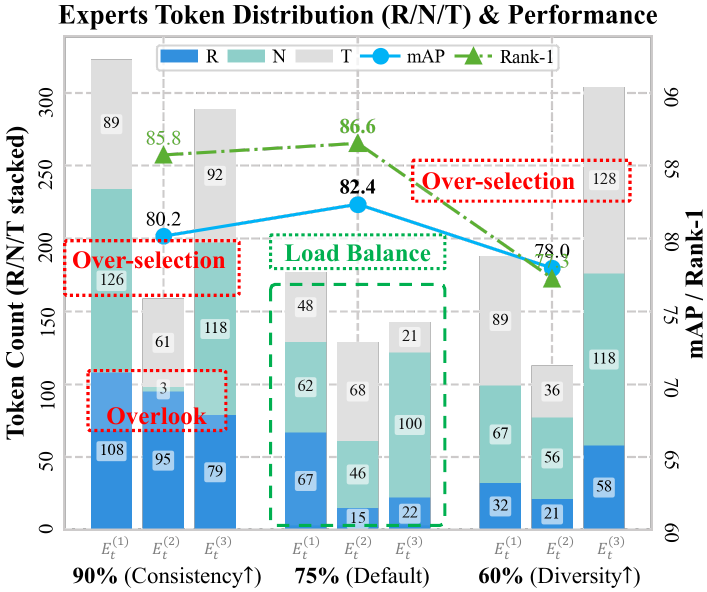}
    \caption{
    Impact of semantic modulation diversity and consistency on expert load balancing and performance. 
    R/N/T denote token counts from RGB, NIR, and TIR modalities for each semantic expert.}
    \label{fig:3_expert_count_distribution}
\end{figure}

\begin{table}
\centering
\renewcommand\arraystretch{1.2}
\caption{Effectiveness of caption type and quality for IDEA and our method.}
\resizebox{0.95\columnwidth}{!}{
\begin{tabular}{c|c||cc|cc}
\noalign{\hrule height 0.8pt}
\multirow{2}{*}{\textbf{Caption Type}} & 
\multirow{2}{*}{\textbf{Quality}} & 
\multicolumn{2}{c|}{\textbf{IDEA}} & 
\multicolumn{2}{c}{\textbf{NEXT (Ours)}} \\
\cline{3-6}
 & & \textbf{mAP} & \textbf{R-1} & \textbf{mAP} & \textbf{R-1} \\
\hline
\hline
IDEA-Text & 100\% & 80.2 & 82.1 & 80.5 & 84.7 \\
\hline
\multirow{3}{*}{NEXT-Text} & 35\%  & 76.1 & 77.9 & 77.1 & 79.7 \\
& 70\%  & 78.1 & 79.9 & 80.0 & 82.2 \\
\rowcolor{table_color}
& \textbf{100\%} & \textbf{80.2} & \textbf{84.0} & \textbf{82.4} & \textbf{86.6} \\
\noalign{\hrule height 0.8pt}
\end{tabular}
}
\label{tb:text_quality}
\end{table}

\begin{table}[t]
\centering
\renewcommand\arraystretch{1.2}
\caption{Effectiveness of sampling strategies.}
\resizebox{0.95\columnwidth}{!}{
\begin{tabular}{c|c||cc|cc}
\noalign{\hrule height 0.8pt}
\multirow{2}{*}{\textbf{Index}} & \multirow{2}{*}{\textbf{Methods}} & \multicolumn{2}{c|}{\textbf{Metrics}}& \multicolumn{1}{c}{\textbf{Params}} & \textbf{Flops}\\
\cline{3-6}
& & \textbf{mAP} & \textbf{R-1} & \textbf{M}& \textbf{G}\\
\hline
\hline
(a) & All-Token & 79.0 & 83.0 & 94.8 & 43.9 \\
(b) & Top-$K$ & 81.4 & 84.3 & 94.8 & 44.8 \\
(c) & Fixed-$\sigma$ & 80.3 & 85.0 & 94.8 & 44.8 \\
\rowcolor{table_color}
(d) & \textbf{$\boldsymbol{\sigma}_{m}$ (Ours)} & \textbf{82.4} & \textbf{86.6} & 94.8 & 44.8\\
\noalign{\hrule height 0.8pt}
\end{tabular}
}
\label{tb:sample_strategy}
\end{table}

\begin{table}[t]
\centering
\caption{Effectiveness of different routing strategies. \textit{Multi.} means each modality uses a different router. \textit{Share.} denotes all modalities use a shared router.}
\renewcommand\arraystretch{1.2}
\resizebox{0.7\columnwidth}{!}{
\begin{tabular}{c||cc|cc}
\noalign{\hrule height 0.8pt}
\multirow{2}{*}{\textbf{Index}} & \multicolumn{2}{c|}{\textbf{Router Type}} & \multicolumn{2}{c}{\textbf{Metrics}}\\
\cline{2-5}
& $\mathbf{E}_{T}$ & $\mathbf{E}_{C}$ & \textbf{mAP} & \textbf{R-1}\\
\hline
\hline
(a) & \textit{Multi.} & \textit{Multi.} & 78.3 & 82.5\\
(b) & \textit{Share.} & \textit{Multi.} & 78.7 & 83.4\\
(c) & \textit{Share.} & \textit{Share.} & 80.3 & 84.3\\
\rowcolor{table_color}
(d) & \textbf{\textit{Multi.}} & \textbf{\textit{Share.}} & \textbf{82.4} & \textbf{86.6}\\
\noalign{\hrule height 0.8pt}
\end{tabular}
}
\label{tb:router_strategy}
\end{table}

\textbf{Semantic Modulation Diversity and Experts Token Distribution.} 
As shown in Fig.~\ref{fig:3_expert_count_distribution}, the sampling ratio in the modulation module plays a crucial role in balancing semantic diversity and consistency.
In Sec.\ref{sec:method}.\textcolor{blue}{B}, the $\textbf{Sampling}$ is a random probability distribution which controls the diversity and consistency of modulated signals.
Intuitively, the sampling probability determines whether a sentence is selected.
A high probability causes the sentence to be sampled frequently, resulting in modulation signals that contain consistent content. 
Conversely, a lower sampling probability produces signals with greater diversity.
A lower probability of 60\% increases semantic diversity but leads to unstable expert routing and a notable decline in performance, reaching 78.0\% mAP and 77.3\% Rank-1.
In contrast, a higher probability of 90\% improves feature consistency but results in slight performance degradation, with 80.2\% mAP and 85.8\% Rank-1, as token over-selection or overlook causes one modality to dominate or suppress in expert activation.
The default probability of 75\% achieves the most desirable trade-off, enabling the model to select semantically balanced tokens across modalities.
Specifically, Experts $E_{t}^{(1)}$, $E_{t}^{(2)}$, and $E_{t}^{(3)}$ select 67, 15, and 22 tokens from RGB, 62, 68, and 100 from NIR, and 48, 68, and 21 from TIR, respectively.
This balanced load preserves modal fairness while maintaining the diversity essential for stable and cooperative expert behavior.

\textbf{Effectiveness of Caption Quality.}
As shown in Table~\ref{tb:text_quality}, we evaluate the sensitivity of caption quality by progressively replacing clean words with noise words.
When adopting 100\% high-quality captions, our method achieves 82.4\% mAP and 86.6\% Rank-1, surpassing IDEA~\cite{wang2025idea} by +2.2\%/+2.6\%.
Replacing captions with those generated by IDEA leads to a 1.9\%/1.9\% drop in mAP and Rank-1, indicating that the richer and more fine-grained semantics in our captions contribute to better visual–text alignment and modality fusion.
As caption quality decreases to 70\%, both models experience noticeable degradation. NEXT drops to 80.0\%/82.2\% mAP and Rank-1, and IDEA falls to 78.1\%/79.9\% mAP and Rank-1, demonstrating the reliance of multi-modal fusion on semantic precision.
When the captions quality is heavily corrupted to 35\%, performance declines sharply, with IDEA and NEXT reducing to 76.1\%/77.9\% and 77.1\%/79.7\% mAP and Rank-1, respectively.
These results indicate that semantic-guided methods are sensitive to the noise in textual captions.
Meanwhile, the captions generated by NEXT provide more effective semantic guidance for multi-model learning, and NEXT has better semantic robustness than IDEA.

\begin{figure}
    \centering
    \includegraphics[width=1.0\linewidth]{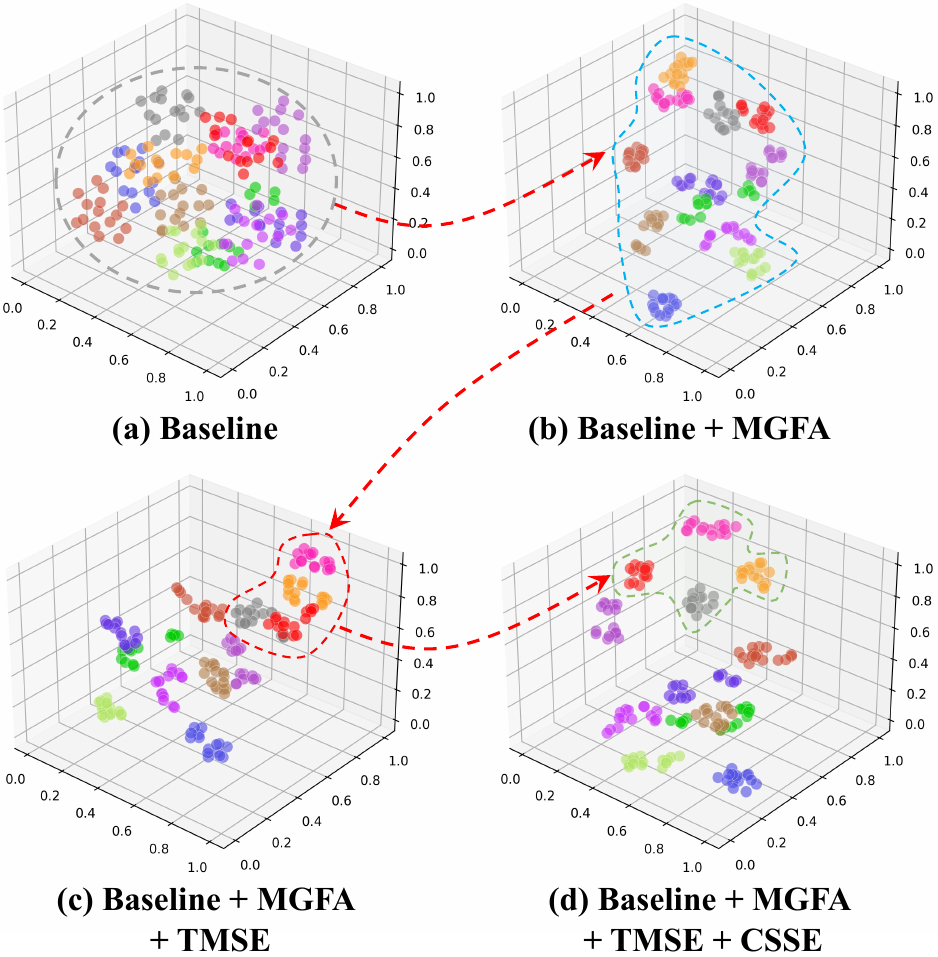}
    \caption{
    T-SNE~\cite{van2008visualizing} visualization of the feature distribution on RGBNT201 (a) Baseline, (b) Baseline + MGFA, (c) Baseline + MGFA + TMSE, and (d) Baseline + MGFA + TMSE + CSSE (Ours).
    }
    \label{fig:4_tsne_201}
\end{figure}

\begin{figure}[t!]
    \centering
    \includegraphics[width=1.0\linewidth]{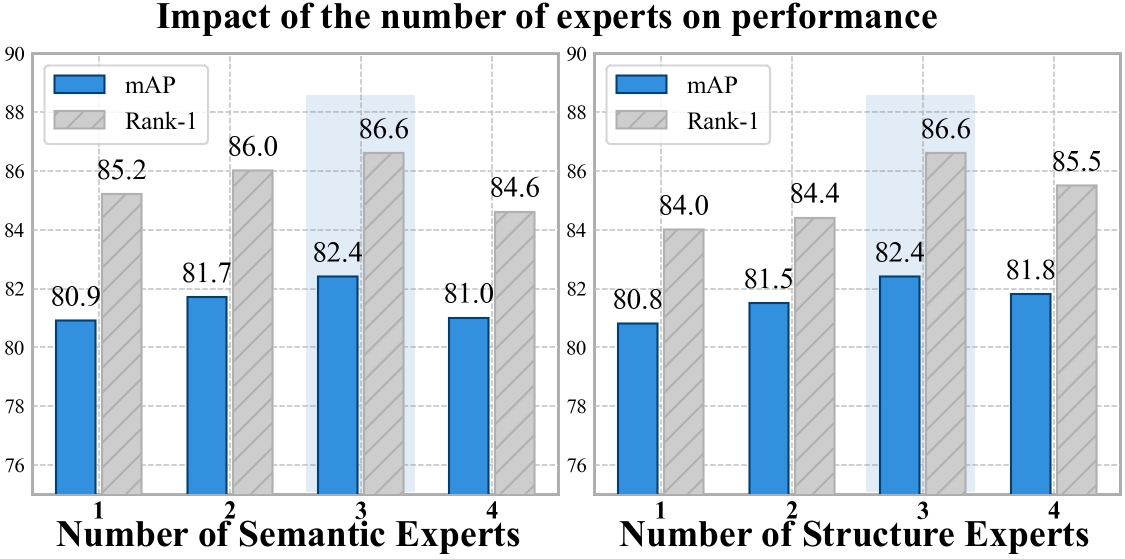}
    \caption{
    Effectiveness of the number of semantic and structural experts.
    }
    \label{fig:2_expert_num_et}
\end{figure}

\begin{figure*}
    \centering
    \includegraphics[width=1.0\linewidth]{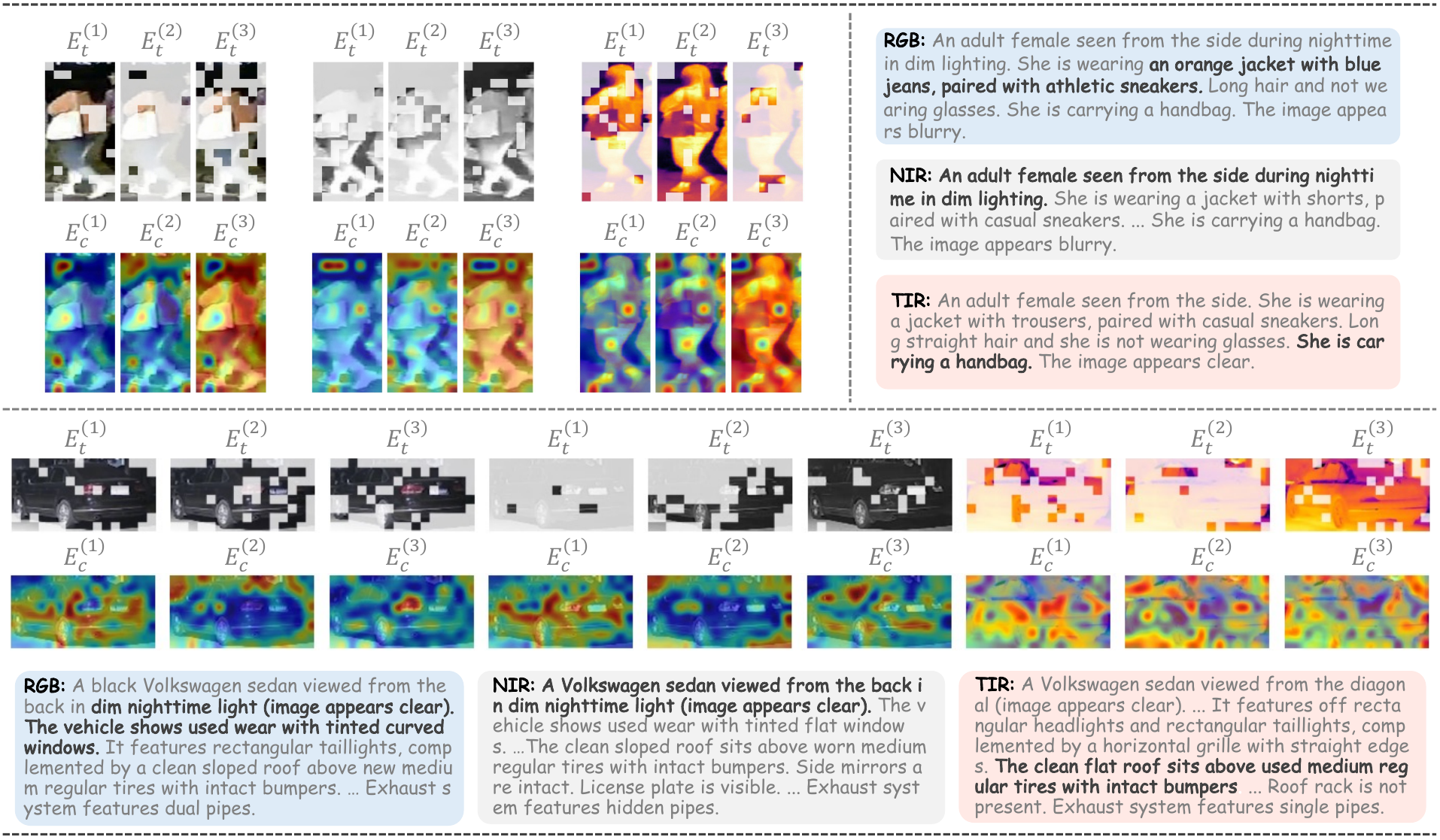}
    \caption{Visualization of the sampled patch tokens and activated feature regions for experts on person and vehicle datasets.}
    \label{fig:3_expert_vis_list2_201_310}
\end{figure*}

\textbf{Effectiveness of Sampling Strategy.}
Table~\ref{tb:sample_strategy} compares different semantic sampling strategies on RGBNT201.
In Table~\ref{tb:sample_strategy}\textcolor{blue}{(a)}, the All-Token strategy encodes all patch tokens $\mathbf{F}_{I,m}^{\mathrm{tok}}$, but fails to identify key semantic regions, resulting in significant performance degradation.
Table~\ref{tb:sample_strategy}\textcolor{blue}{(b)} adopts the Top-$K$ strategy, replacing the dynamic routing threshold $\boldsymbol{\sigma}_{m}$ in $\mathbf{R}_{m}$ with the top 50\% tokens, achieving a suboptimal result.
Table~\ref{tb:sample_strategy}\textcolor{blue}{(c)} replaces the $\boldsymbol{\sigma}_{m}$ with a Fixed-$\sigma$, which lacks adaptive perception and relies on handcrafted tuning, leading to inferior performance.
In contrast, our dynamic sampling strategy achieves optimal performance, confirming its effectiveness.

\textbf{Effectiveness of Routing Strategy.}
As shown in Table~\ref{tb:router_strategy}, different routing strategies are explored to investigate the impact of modality-specific or shared routing on performace.
When both the semantic expert $\mathbf{E}_T$ and the structural expert $\mathbf{E}_C$ adopt modality-specific routing in Table~\ref{tb:router_strategy}\textcolor{blue}{(a)}, the model achieves 78.3\% mAP and 82.5\% Rank-1.
Sharing the routing for $\mathbf{E}_T$ slightly improves performance in Table~\ref{tb:router_strategy}\textcolor{blue}{(b)}, while using shared routing for both experts in Table~\ref{tb:router_strategy}\textcolor{blue}{(c)} further boosts the results to 80.3\% mAP and 84.3\% Rank-1, suggesting that a certain level of cross-modal coupling is beneficial.
The optimal configuration in Table~\ref{tb:router_strategy}\textcolor{blue}{(d)} is achieved when $\mathbf{E}_T$ uses modality-specific routers and $\mathbf{E}_C$ dopts a shared router, reaching 82.4\% mAP and 86.6\% Rank-1. 
Intuitively, semantic cues such as color, illumination, and temperature are modality-dependent and thus should be learned via modality-specific routing, while structural cues (\eg, object shape, contour, or posture) are modality-invariant, making shared routing more suitable for consistent structural perception.

\textbf{Effectiveness of Experts Number.}
As shown in Fig.~\ref{fig:2_expert_num_et}, the impact of the number of experts on model performance is analyzed.
Increasing the number of semantic experts from one to three progressively improves performance, with mAP rising from 80.9\% to 82.4\% and Rank-1 increasing from 85.2\% to 86.6\%. 
Introducing a fourth expert leads to a decline in both metrics, dropping to 81.0\% in mAP and 84.6\% in Rank-1. A similar trend is observed for structural experts. 
Expanding the expert count from one to three steadily boosts performance, with mAP growing from 80.8\% to 82.4\% and Rank-1 from 84.0\% to 86.6\%, while four experts introduce a clear performance drop. 
These results indicate that an excessive number of experts increases redundancy and weakens specialization, whereas three experts reach an effective balance between representational richness and expert diversity, yielding the most discriminative identity features.

\begin{figure}
    \centering
    \includegraphics[width=1.0\linewidth]{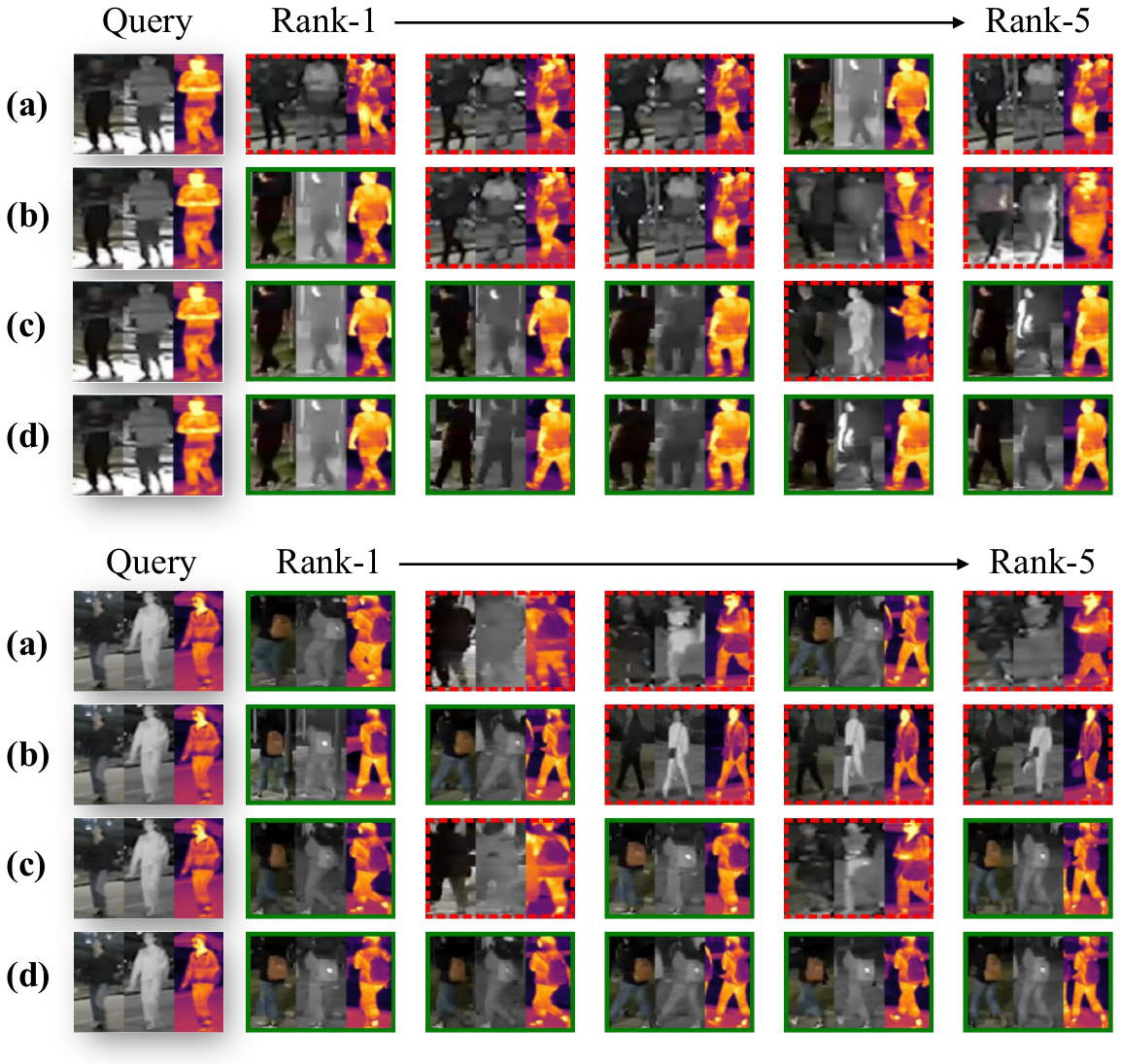}
    \caption{ 
    Comparison of retrieval results for different modules on RGBNT201: (a) Baseline, (b) Baseline + MGFA, (c) Baseline + MGFA + TMSE, and (d) Baseline + MGFA + TMSE + CSSE (Ours).
    }
    \label{fig:5_rank_list_201_modal_short}
\end{figure}

\begin{table}[t]
\centering
\renewcommand\arraystretch{1.2}
\caption{Comparison with state-of-the-art methods on the size of learnable parameters, flops, and throughput on RGBNT201. $^*$ denotes the use of CLIP text encoder in the inference stage.}
\resizebox{1.0\columnwidth}{!}{
\begin{tabular}{l||ccc|cc}
\noalign{\hrule height 0.8pt}
\multirow{2}{*}{\textbf{Methods}}
& \textbf{Params} & \textbf{FLOPs} & \textbf{Throughput} & \multicolumn{2}{c}{\textbf{Metrics}} \\
\cline{2-6}
& \textbf{M} & \textbf{G} & \textbf{Images/s} & \textbf{mAP} & \textbf{R-1}\\
\hline
\hline
\text{EDITOR}~\cite{zhang2024magic} & 119.3 & 40.8 & 335.1 & 66.5 & 68.3 \\
\text{TOP-ReID}~\cite{DBLP:conf/aaai/WangLZLTL24} & 324.5 & 35.5 & 398.9 & 72.3 & 76.6 \\
\text{ICPL-ReID}~\cite{li2025icpl} & 45.4 & 39.8 & 358.4 & 75.1 & 77.4 \\
\text{PromptMA}~\cite{10955143} & 107.9 & 36.2 & 343.5 & 78.4 & 80.9 \\
\text{DeMo}~\cite{wang2024demo} & 98.8 & 35.1 & 403.6 & 79.0 & 82.3 \\
\text{MambaPro}~\cite{Wang2024MambaPro} & 74.8 & 52.4 & 243.2 & 78.9 & 83.4 \\
\rowcolor{table_color}
\textbf{NEXT (w/o Text)} & 94.8 & 36.0 & 354.0 & 79.2 & 85.5 \\
\hline
\text{IDEA$^*$}~\cite{wang2025idea} & 91.7 & 43.7 & 299.5 & 80.2 & 82.1 \\
\rowcolor{table_color}
\textbf{NEXT$^*$ (Ours)} & 94.8 & 44.8 & 275.1 & \textbf{82.4} & \textbf{86.6} \\
\noalign{\hrule height 0.8pt}
\end{tabular}
}
\label{tb:params_flops}
\end{table}

\textbf{Training and Inference Efficiency Analysis.}
Table~\ref{tb:params_flops} compares the efficiency of various methods on the RGBNT201 dataset in terms of learnable parameters, flops, throughput, and metrics accuracy. 
Our method NEXT achieves the best performance 82.4\%/86.6\% mAP and Rank-1, while maintaining a moderate model size of 94.8M parameters.
Compared to the MoE-based method DeMo~\cite{wang2024demo} (98.8M), NEXT reduces the parameter count by 4M, demonstrating the higher efficiency. 
While adding the text encoder increases the FLOPs of NEXT by 8.8 relative to the version without text, it still maintains a clear computational advantage over MambaPro~\cite{Wang2024MambaPro}, which consumes 52.4G.
For inference throughput, NEXT achieves 275.1 Images/s, which is slightly lower than IDEA's 299.5 Images/s, but outperforms MambaPro‘s 243.2 Images/s.
These results validate that NEXT obtains superior performance with a trade-off between model complexity and efficiency.

\begin{figure}
    \centering
    \includegraphics[width=1.0\linewidth]{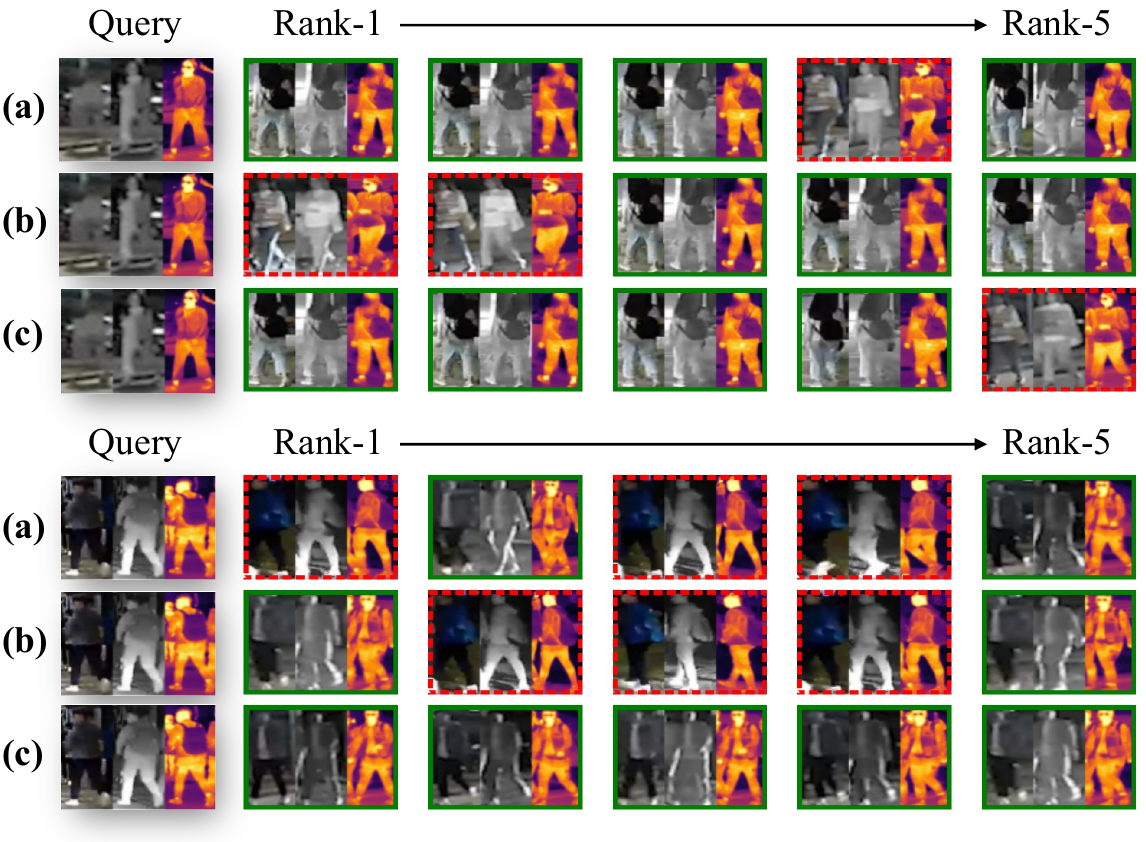}
    \caption{  
    Comparison of retrieval results for state-of-the-art methods on RGBNT201: (a) DeMo, (b) IDEA, and (c) NEXT (Ours).
    }
    \label{fig:5_rank_list_201_demo_idea_next_short}
\end{figure}

\subsection{Visualization Analysis}
\textbf{Feature Distribution.}
As shown in Fig.~\ref{fig:4_tsne_201}, we utilize the T-SNE~\cite{van2008visualizing} to visualize the feature distributions on RGBNT201.
Comparing Figs.~\ref{fig:4_tsne_201}\textcolor{blue}{(a)} and \textcolor{blue}{(b)}, the introduction of MGFA effectively aggregates different instances of the same identity.
In Fig.~\ref{fig:4_tsne_201}\textcolor{blue}{(c)}, incorporating the TMSE semantic experts improves the separability of samples across identities.
In Fig.~\ref{fig:4_tsne_201}\textcolor{blue}{(d)}, the inclusion of CSSE structural experts leads to tighter clustering of hard samples, resulting in well-separated intra-class and inter-class distributions.
These results validate the effectiveness of each proposed module.

\textbf{Expert Feature Visualization.} 
As shown in Fig.~\ref{fig:3_expert_vis_list2_201_310}, we visualize the sampling masks of the semantic experts and the activation maps of the structural experts. 
We observe that different semantic experts $E^{(i)}_{t}$ exhibit distinct modality preferences: for instance in person dataset, $E^{(1)}_{t}$ tends to sample features from the RGB modality, $E^{(2)}_{t}$ from the TIR modality, and $E^{(3)}_{t}$ from the NIR modality.
These sampling patterns are notably complementary across modalities, indicating that the text-modulated expert effectively learns complementary modality-specific features to enhance multi-modal representations.
On the other hand, structural experts maintain the integrity of coarse-grained identity structures and environmental context awareness. 
The parts ignored by expert $E^{(3)}_{c}$ are well captured by other structural experts. 
These observations validate that NEXT successfully decouples identity recognition into semantic sampling and structure modeling. 

\textbf{Retrieval Results.}
To verify the model’s effectiveness in real world scenarios, we visualize the retrieval results of models with different module combinations on the RGBNT201 dataset and compare NEXT with state-of-the-art methods.
As shown in Fig.~\ref{fig:5_rank_list_201_modal_short}, the introduction of different modules enables the model to recognize pedestrians with the same identity more effectively.
Fig.~\ref{fig:5_rank_list_201_demo_idea_next_short} compares our method with existing state-of-the-art methods, demonstrating that our method achieves superior retrieval performance for the target person under various low-light conditions.
These results validate the effectiveness of NEXT in leveraging multi-modal data and textual semantics for robust person ReID.

\section{Conclusion}
In this paper, we focus on the explicit semantic learning of multi-modal object ReID.
We first propose a reliable multi-modal caption generation pipeline based on MLLMs, which utilize template-based instructions to construct confidence-aware attribute sets, and merge multi-modal attributes to compose high-quality captions.
Additionally, we propose the multi-modal ReID network, NEXT, a novel multi-grained MoE framework. 
Specifically, we propose the Text-Modulated Semantic Experts (TMSE) and the Context-Shared Structure Experts (CSSE) to decouple the recognition problem into fine-grained semantic sampling and coarse-grained structural perception.
We incorporate a Multi-Grained Features Aggregation (MGFA) module to unify features from multi-grained experts and obtain complete identity representations.
Extensive experiments on five popular datasets demonstrate the effectiveness of our method.
In the future, we plan to further explore the reasoning capability of MLLMs to drive deeper semantic fusion, such as harder negative mining and chain-of-thought reasoning.
We believe this direction will unlock new frontiers in interpretable and semantic-driven multi-modal ReID.

\bibliographystyle{IEEEtran}
\bibliography{ieee_next}{}

\newpage

 
\vspace{11pt}




\vfill

\end{document}